\documentclass[sigconf]{aamas}

\usepackage{balance} 
\usepackage{algorithm}
\usepackage{algorithmic}
\usepackage{amsmath}
\usepackage{longtable}
\usepackage{booktabs}
\usepackage{pifont}
\usepackage{multicol}
\usepackage{multirow}

\usepackage[utf8]{inputenc}
\usepackage[most]{tcolorbox}
\tcbuselibrary{listingsutf8}

\doi{SBAY6258}

\makeatletter
\gdef\@copyrightpermission{
  \begin{minipage}{0.2\columnwidth}
   \href{https://creativecommons.org/licenses/by/4.0/}{\includegraphics[width=0.90\textwidth]{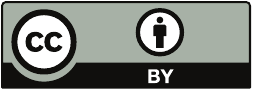}}
  \end{minipage}\hfill
  \begin{minipage}{0.8\columnwidth}
   \href{https://creativecommons.org/licenses/by/4.0/}{This work is licensed under a Creative Commons Attribution International 4.0 License.}
  \end{minipage}
  \vspace{5pt}
}
\makeatother

\setcopyright{ifaamas}
\acmConference[AAMAS '26]{Proc.\@ of the 25th International Conference
on Autonomous Agents and Multiagent Systems (AAMAS 2026)}{May 25 -- 29, 2026}
{Paphos, Cyprus}{C.~Amato, L.~Dennis, V.~Mascardi, J.~Thangarajah (eds.)}
\copyrightyear{2026}
\acmYear{2026}
\acmDOI{}
\acmPrice{}
\acmISBN{}

\acmSubmissionID{1634}

\title{DRAGON: LLM-Driven Decomposition and Reconstruction Agents for Large-Scale Combinatorial Optimization}

\author{Shengkai Chen}
\affiliation{
  \institution{Institute for Infocomm Research}
  \city{A*STAR}
  \country{Singapore}}
\email{Chen_Shengkai@a-star.edu.sg}

\author{Zhiguang Cao}
\affiliation{
  \institution{Singapore Management University}
  \country{Singapore}}
  \email{zgcao@smu.edu.sg}

\author{Jianan Zhou}
\affiliation{
  \institution{Nanyang Technological University}
  \country{Singapore}}
\email{jianan004@e.ntu.edu.sg}

\author{Yaoxin Wu}
\affiliation{
  \institution{Eindhoven University of Technology}
  \city{Eindhoven}
  \country{Netherlands}}
\email{y.wu2@tue.nl}

\author{Senthilnath Jayavelu}
\affiliation{
  \institution{National University of Singapore \& Institute for Infocomm Research}
  \city{A*STAR}
  \country{Singapore}
  }
\email{J_Senthilnath@a-star.edu.sg}

\author{Zhuoyi Lin}
\authornote{Corresponding author}
\affiliation{
  \institution{Institute for Infocomm Research}
  \city{A*STAR}
  \country{Singapore}
  }
\email{Lin_Zhuoyi@a-star.edu.sg}

\author{Xiaoli Li}
\affiliation{
  \institution{Singapore University of Technology}
  \city{and Design}
  \country{Singapore}}
\email{xiaoli_li@sutd.edu.sg}

\author{Shili Xiang}
\affiliation{
  \institution{Institute for Infocomm Research}
  \city{A*STAR}
  \country{Singapore}}
\email{Xiang_Shili@a-star.edu.sg}

\begin{abstract}
Large Language Models (LLMs) have recently shown promise in addressing combinatorial optimization problems (COPs) through prompt-based strategies. However, their scalability and generalization remain limited, and their effectiveness diminishes as problem size increases, particularly in routing problems involving more than 30 nodes. 
We propose \textbf{DRAGON}, which stands for \textbf{D}ecomposition and \textbf{R}econstruction \textbf{A}gents \textbf{G}uided \textbf{O}ptimizatio\textbf{N}, a novel framework that combines the strengths of metaheuristic design and LLM reasoning.
DRAGON autonomously identifies regions in initial solution with high optimization potential and strategically decompose large-scale COPs into manageable subproblems. Each subproblem is then reformulated as a concise, localized optimization task and solved through targeted LLM prompting guided by accumulated experiences. Finally, the locally optimized solutions are systematically reintegrated into the global context to yield a significantly improved outcome. 
By continuously interacting with the optimization environment and experience memory, the agents iteratively learn from feedback, effectively coupling symbolic reasoning with heuristic search.
Empirical results show that, DRAGON consistently produces feasible solutions on TSPLIB, CVRPLIB, and Weibull-5k bin packing benchmarks, and achieves near-optimal results (0.16\% gap) on knapsack problems with over 3M variables. This work shows the potential of feedback-driven language agents as a new paradigm for generalizable and interpretable large-scale optimization.
\end{abstract}


\newcommand{\BibTeX}{\rm B\kern-.05em{\sc i\kern-.025em b}\kern-.08em\TeX}

\begin{document}


\pagestyle{fancy}
\fancyhead{}


\maketitle 


\section{Introduction}
\begin{figure}[!t]
    \centering
    \includegraphics[width=0.9\linewidth]{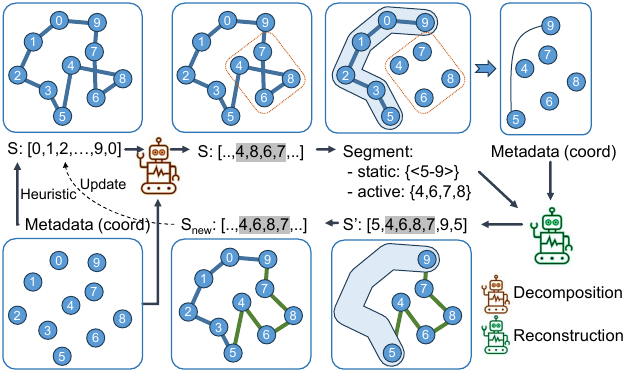}
    \caption{Illustration of DRAGON on a TSP instance. The decomposer identifies a suboptimal segment \{4,6,7,8\} (grayed) from a global solution. The segment is locally refined and reintegrated by the reconstructor to improve the global tour.}
    \label{fig:illustrate_tsp}
\end{figure}

Combinatorial optimization problems (COPs), such as the Traveling Salesman Problem (TSP), Vehicle Routing Problem (VRP), and Knapsack Problem (KP), are notoriously challenging due to the NP-hard nature \cite{ausiello2012complexity}. COPs are typically addressed using exact algorithms or heuristic methods, which often require substantial manual design and parameter tuning \cite{rokbani2021bi,cacchianiKnapsackProblemsOverview2022,ma2023learning, lin2024cross, feng2025lifelong}, thus limiting scalability and adaptability to varying problem instances and domains.

Large Language Models (LLMs) have led to a series of breakthroughs for various challenging problems, including language understanding, text generation, code synthesis, and reasoning tasks \citep{brown2020language, chowdhery2023palm, achiam2023gpt}.
Recent evidence reveals that LLMs have emerged as powerful tools capable of addressing a variety of COPs, leveraging their reasoning capabilities for direct solution generation \cite{yang2024large, wang2024can} or heuristic design \cite{ye2024reevo, romera2024mathematical, liu2024evolution, zhu2025bridging}.
Their inherent strengths in abstraction, generalization, and semantic understanding allow them to generate promising solutions even without explicit algorithmic instructions. 
However, the current capabilities of LLMs for directly generating solutions remain largely confined to relatively small-scale instances, such as TSP with fewer than 30 nodes \cite{yang2024large, iklassov2024self, liu2024large}.
As problem size and complexity increase, LLM-based solutions often deteriorate due to restricted context length, reduced logical coherence, and difficulty in representing combinatorial structures \cite{iklassov2024self, zhao2025can}. These limitations significantly impede the practical deployment of LLM-driven methods in real large-scale applications, such as logistics, transportation, and supply chain management, where problems commonly involve hundreds to thousands of nodes \cite{bengio2021machine}. 



In this study, we aim to answer two research questions: 
(1) \emph{Can an LLM agent, without solving the full COP, identify and isolate regions of the solution that are likely suboptimal or have improvement potential?}
(2) \emph{Can an LLM agent effectively solve small-scale COPs locally while following additional, customized non-trivial constraints?}
These questions highlight the core challenges of our framework: utilizing LLMs to guide decomposition by detecting potential improvement areas, and ensuring reconstructed solutions remain feasible.

Prompt-based LLM methods often fail on large-scale instances due to the absence of domain knowledge and the context length limitation, while code-generation approaches require extensive task-specific training to develop robust heuristics, leading to substantial computational costs and poor cross-domain transferability. 
Meanwhile, classical metaheuristics such as divide-and-conquer and large neighborhood search (LNS) \cite{pisinger2018large} exhibit strong scalability in large-scale COPs by iteratively decomposing and refining solutions, though they rely heavily on handcrafted heuristics and expert knowledge. 
To address the gap, we propose \textbf{DRAGON}, a framework that transforms complex large COP solving into decomposition and reconstruction tasks guided by LLM agents.
DRAGON first strategically identifies regions with high potential of improvement, decomposing large-scale COPs into context-manageable subproblems, and locally optimizes each subregion through reconstruction. 

To concretely illustrate this idea, Figure~\ref{fig:illustrate_tsp} demonstrates an example of how the proposed Dragon improves TSP solution locally and globally. Given an initial global solution, the decomposition agent $\mathcal{D}$ identifies nodes $\{4,6,7,8\}$ (highlighted in gray) as a subproblem with high potential for further improvement, while the remaining nodes remain static and impose boundary constraints. 
During reconstruction, the suboptimal local segment (4-8-6-7) is locally optimized with LLM agents, leading to a better sub-solution (4-6-7-8). Consequently, the improved local segment is integrated into the global solution, yielding a reduced tour length solution.

Overall, our contributions are threefold:
(1) 
To the best of our knowledge, this is the first work to demonstrate that LLM agents can be effectively leveraged to directly generate high-quality solutions for large-scale COPs, opening new possibilities for LLM-driven optimization.
(2) We propose DRAGON, a divide-and-conquer framework that dynamically decomposes large-scale COPs and subsequently refines compact subproblems by state passing between agents. 
(3) We empirically validate DRAGON on large-scale COPs benchmarks, demonstrating substantial improvements in solution quality and scalability compared to state-of-the-art LLM-based baselines. 
Our findings validate the feasibility of leveraging LLM agents for large-scale COPs that are prevalent in real-world applications.


\section{Related Work}
\textbf{LLMs for Combinatorial Optimization.} Recent studies have increasingly examined the capability of LLMs to solve COPs by leveraging prompt engineering techniques.
Among these efforts, a prominent direction involves using LLMs directly as solution generators, where carefully designed prompts and enriched input information guide the models to produce viable solutions.
Early research by \citet{wang2024can} provided evidence that LLMs could solve certain graph-based COPs, although this capability was strongly tied to the quality of the provided prompts. Subsequent work has sought to refine this process by incorporating contextual information. Examples include integrating graph structural and topological data \cite{wasserkrug2024large}, existing heuristic solutions \cite{iklassov2024self}, explicit relationships between solutions and objectives \cite{yang2024large}, and visual representations of problems or solutions \cite{huang2024multimodal, elhenawy2024eyeballing}.
Another research pathway involves leveraging LLMs as components within structured algorithms or frameworks. For instance, \citet{liu2024large} demonstrated the utility of LLMs in evolutionary computation by prompting them to perform essential algorithmic operations such as selection, crossover, and mutation. Similarly, \citet{elhenawy2024visual} deployed multiple LLMs as collaborative agents, each responsible for distinct roles such as initialization, critique, and evaluation within a predefined optimization pipeline.
Despite these promising advancements, current capabilities for directly generating solutions via LLMs remain largely restricted to relatively small-scale problems, typically limited to instances such as TSP with fewer than 30 nodes. To circumvent this limitation, \citet{iklassov2024self} introduced a decomposition approach wherein an LLM assesses problem complexity autonomously. If deemed difficult, the model recursively subdivides the problem into simpler, manageable subproblems. Nonetheless, its scalability remains limited.

Beyond direct solution generation, other studies have explored leveraging LLMs for heuristic design \cite{romera2024mathematical, liu2024evolution, ye2024reevo} and mathematical modeling \cite{xiao2023chain, ahmaditeshnizioptimus,jiang2025droc}. The former aims to use LLMs to discover heuristics expressed in code that have potential to outperform those crafted by human experts, while the latter focuses on translating natural language problem descriptions into mathematical formulations compatible with traditional OR solvers. More recently, research has investigated training large language models for end-to-end combinatorial optimization, rather than relying solely on prompting paradigms~\cite{jianglarge,jiang2024bridge}. These directions fall outside the scope of this work, and we refer interested readers to \citet{da2025large} for a comprehensive survey.

\noindent\textbf{Large-Scale Combinatorial Optimization.}  Another research direction related to this work aiming to design metaheuristics for solving large-scale COPs, either with machine learning \cite{li2021learning,sun2023difusco} or through traditional methods \cite{shaw1998using,arnold2019efficiently}. These approaches generally follow decomposition or divide-and-conquer principles. While they have shown promising results, their effectiveness often relies on problem-specific strategies (e.g., domain-specific decomposition rules), which limits their generality. In this work, we propose DRAGON, a general decomposition-reconstruction framework leveraging the semantic capabilities of LLMs. Unlike previous methods, DRAGON identifies suboptimal regions without explicit domain-specific rules and iteratively reconstructs globally consistent solutions. Our framework effectively addresses LLM context-length constraints, combining learned decomposition advantages with the generalization provided by LLM agents.

\section{Methodology}

\begin{figure*}[t]
    \centering
    \includegraphics[width=0.9\linewidth]{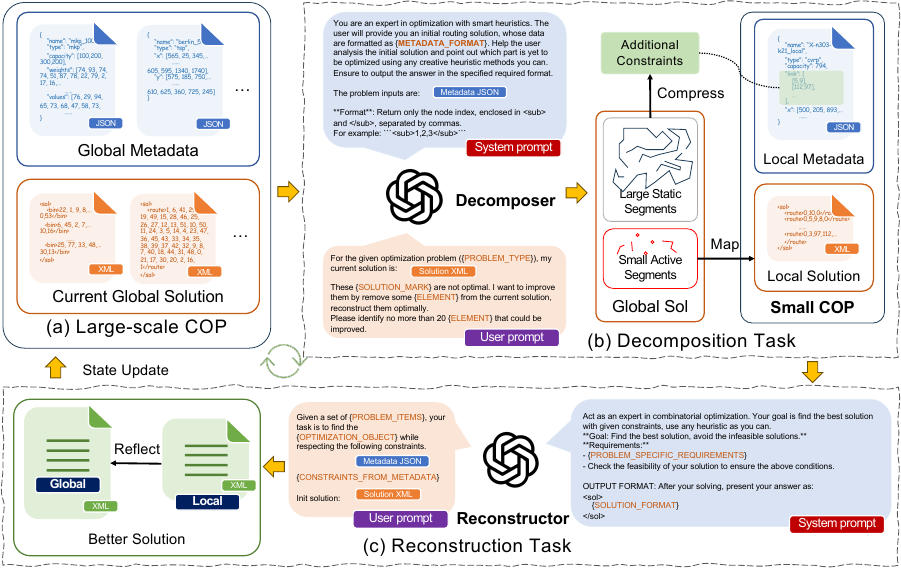}
    \caption{Overview of the state passing among agents in DRAGON framework. With (a) given COP data input and current global solution, the DRAGON pipeline process alternates between two key stages: (b) A decomposition step uses an LLM to split a large-scale COP into manageable active and static segments. These are compressed into a smaller subproblem; (c) A reconstruction step solves the reduced problem with additional constraints to yield a refined local solution.}
    \label{fig:framework}
\end{figure*}


DRAGON enables LLM agents to effectively solve large-scale COPs via state passing communication. As illustrated in Figure~\ref{fig:framework}, DRAGON iteratively alternates between two core tasks: Decomposition and Reconstruction. Starting from fast heuristic initialization, DRAGON progressively refines it by identifying regions with high improvement potential and solving localized subproblems under explicit constraint awareness. In each iteration, the current solution is decomposed into manageable subproblems, locally optimized through reconstruction agents, and then reassembled into a globally improvement. This hierarchical process allows DRAGON to scale from small instances to complex large-scale COPs. 
Algorithm~\ref{alg:workflow} summarized our workflow, where implicit communication between agents occurs via solution state passing.

\begin{algorithm}[th]
\footnotesize
\caption{DRAGON for Large-Scale COPs}
\label{alg:workflow}
\begin{algorithmic}[1]
\REQUIRE Metadata $M$, initial solution $S_0$, maximum time $T_{\max}$, rejection threshold $N$
\ENSURE Improved solution $S$

\STATE $S \gets S_0$ \hfill // Current solution
\STATE $t \gets 0$ \hfill // Time counter
\STATE $r \gets 0$ \hfill // Rejection counter
\WHILE{$t < T_{\max}$ \AND $r < N$}
    \STATE $(\mathbf a, \mathbf s) \gets \mathcal{D}(M, S)$ \hfill // \textbf{Decomposition}
    \STATE $(M',S', C) \gets \texttt{Compress}(M, \mathbf a, \mathbf s)$ \hfill 
    \STATE $S'_{\text{new}} \gets \mathcal{R} (M', S', C)$ \hfill // \textbf{Reconstruction}
    \IF{\texttt{Accept}$(S', S'_{\text{new}})$}
        \STATE $S' \gets S'_{\text{new}}$
        \STATE $S \gets \texttt{Integration}(S',S)$
        \STATE $r \gets 0$ \hfill // Reset rejection counter
    \ELSE
        \STATE $r \gets r + 1$
    \ENDIF
    \STATE $t \gets \texttt{ElapsedTime}()$ \hfill // Reach running time limit.
\ENDWHILE
\RETURN $S$
\end{algorithmic}
\end{algorithm}




\subsection{Framework Inputs and Solution Updates}
DRAGON interacts with problem environment that contains two fundamental inputs: metadata $M$ and the current solution $S$. Metadata $M$, represented as structured JSON, encodes essential problem-specific parameters, and the solution $S$ is represented as an ordered sequence of elements, such as city indices for TSP or item identifiers for Knapsack, and is formatted in either JSON or XML, as illustrated in Figure~\ref{fig:framework}(a).

Formally, given metadata $M$, the global solution at iteration $i$ is obtained as follows:
\begin{equation}
    S_i = \begin{cases}
    \texttt{Fast\_Heuristic}(M), & i=0 \\
    \texttt{Integration}(S'_{i-1}, S_{i-1}), & i>0 \\
    S'_{i-1} = \mathcal R \left(\mathcal D\left(M,S_{i-1}\right)\right), & i>0 
    \end{cases}
\end{equation}
Here, $S'_{i-1}$ denotes the locally refined sub-solution from the previous iteration, identified by the decomposer $\mathcal{D}$ and enhanced by the reconstructor $\mathcal{R}$ (both detailed in subsequent sections). \texttt{Integration} involves simply concatenation and sorting operations, while specific implementation of \texttt{Fast\_Heuristic} are provided in the experimental section.
Furthermore, a new solution is conditionally accepted using a probabilistic criterion (assuming minimization):
\begin{equation}
    \texttt{Accept}(S_1, S_2) = \min\left\{1, \exp\left(-\frac{f(S_2) - f(S_1)}{T}\right)\right\}\label{eq:accept}
\end{equation}
where $f$ denotes the objective function and $T$ is a temperature parameter controlling the acceptance probability. As shown in Algorithm~\ref{alg:workflow}, we substitute the current local solution $S'_i$ and its $\mathcal{R}$ processed version $S'_{\text{new},i}$ into Eq.~\ref{eq:accept} as $S_1$  and $S_2$, respectively, and accepted solutions are then integrated back into the global solution to ensure consistency and feasibility.
It is worth noting that even when the new local solution is worse i.e., $f(S'_{\text{new},i}) \ge f(S'_i)$, there remains a small chance of acceptance and integration, allowing DRAGON to occasionally accept inferior solutions, helping it escape local optima, promote exploration of the solution space, and maintain a balancing between exploitation and exploration.

Balancing exploitation (greedy improvement) and exploration (diversification) is critical to the performance of heuristic search methods. In many cases, domain experts must carefully design problem-specific strategies to achieve this balance, and the convergence rate of the solution gap can vary significantly depending on the designer’s expertise. However, for large-scale COPs where analyzing the problem structure is extremely challenging and generalization across domains is difficult, we leverage LLM-based agents, which are well-suited for both exploitation and exploration, because they can exploit stored experiences from previous trails, while their generative and adaptive capabilities enable diverse solution proposals, supporting broader exploration of the solution space.

\subsection{Decomposition for Large-Scale COPs}
A central challenge in applying LLMs to large-scale COP decomposition lies in their limited capability to reason over complex global constraints. Although existing studies demonstrate the potential of LLMs in solving COPs~\cite{wilson2023,yang2024large}, their capability to accurately detect and isolate sub-problems suitable for local refinement within large-scale COPs remains largely unexplored. 
To this end, we propose a decomposition agent ($\mathcal{D}$) which leverages LLMs to strategically partition COPs into smaller, manageable sub-problems for further refinement, rather than directly solving the entire COP. 
Specifically, the decomposer $\mathcal{D}$ functions as a high-level planner, analyzing metadata $M$ and the current solution $S_i$. It partitions the solution into \emph{active segments} ($\mathbf{a}_i$), which have substantial improvement potential, and \emph{static segments} ($\mathbf{s}_i$), which remain unchanged during the current iteration. Specifically, decomposition is formulated as:
\begin{equation}
(\mathbf a_i, \mathbf s_i) = \mathcal{D}(M, S_i)
\end{equation}


DRAGON demonstrates that well-designed prompts (illustrated in Figure~\ref{fig:framework}(b) and Supplementary Materials (Prompt designs)) enable LLMs to effectively identify segments with significant potential for improvement while introducing additional constraints ($C_i$) to maintain feasibility and global consistency:
\begin{equation}
(M'_i, S'_i, C_i) = \texttt{Compress}(M,\mathbf{a}_i, \mathbf{s}_i)
\end{equation}
where $M'_i \subseteq M$ denotes the subset of original metadata relevant only to the active segments.
For example, in the TSP scenario illustrated in Figure~\ref{fig:illustrate_tsp}, the decomposer identifies an active segment $\mathbf{a}_i = \{4,6,7,8\}$ and a static segment $\mathbf{s}_i = [9,0,1,2,3,5]$. The \texttt{Compress} step significantly reduces the size of the local data $|M_i'| << |M|$, making it manageable to downstream agents.
However, it will introduce extra specific constraints, which are a set of compressed representations of static segment $\mathbf s_i$.

\subsection{Reconstruction with Constraint Integration}
Recent studies ~\cite{yang2024large,liu2024large,iklassov2024self} highlight the effectiveness of LLMs in solving small-scale COPs, typically with fewer than 30 decision variables. Building on this capability, our reconstruction agent ($\mathcal{R}$) is designed to solve compressed COP instances iteratively, derived from the decomposition stage.

\begin{figure}[t]
    \centering
    \includegraphics[width=\linewidth]{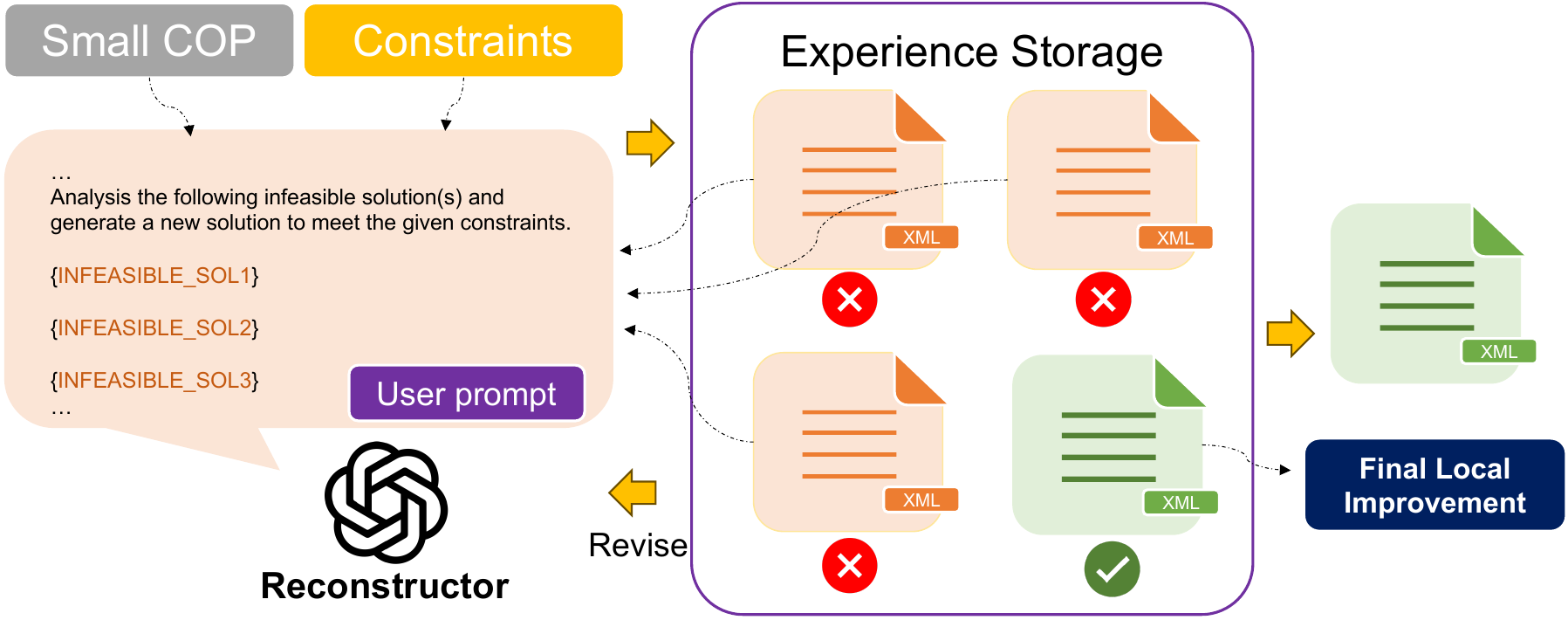}
    \caption{Constraint satisfaction in reconstruction is ensured via experience accumulation.}
    \label{fig:revise}
\end{figure}

Specifically, reconstructor $\mathcal{R}$ utilizes localized metadata $M'_i$ augmented with explicit constraints $C_i$ to ensure global feasibility and consistency. Constraints $C_i$ are explicitly integrated into natural language within the LLM prompt, with the formats varying by COP type. Formally, the reconstruction process is defined as: 
\begin{equation}
S'_{\text{new},i} = \mathcal{R}(M'_i, S'_i, C_i)
\end{equation}
where $M'_i$ encapsulates localized metadata, and $C_i$ encodes global consistency requirements derived from static segments $\mathbf{s}_i$.


Note that the format and interpretation of constraints depend on the specific COP type. In routing problems such as TSP and CVRP, constraints typically enforce particular edge sequences to ensure route continuity, while in packing problems like MKP or BPP, constraints explicitly mandate item inclusion or exclusion. For instance, a constraint prompt for TSP may state: \textit{``the following edges must be visited consecutively, with no additional points permitted between them.''} It is essential to maintain consistency between constraints and compressed metadata. 
For example, required edges $(a,b)$ in CVRP represent condensed segments from the original route, with associated demands accurately reflecting the cumulative demands of intermediate nodes.


As illustrated in Figure~\ref{fig:revise},  reconstruction agent utilizes iterative experience storage collected along previous trails to self-revise candidate solutions until all specified constraints are satisfied. The agent's experience storage records all previous attempts, regardless of whether they result in feasible solutions or not, along with annotations on the type and cause of any constraint violations.
This memory allows the agent to learn from past infeasible solutions, avoid repeating similar errors, and identify patterns within feasible solutions to guide future improvements. Through this process of iterative refinement, localized solutions are gradually adjusted to remain coherent and globally feasible. Extensive experiments detailed in the experimental section and Supplementary Materials demonstrate that, when guided by well-structured prompts and clearly defined constraints, LLM agents can consistently generate high-quality solutions to constrained subproblems.


\section{Experiments}

\subsection{Setup}
\subsubsection{Problem Descriptions}
To evaluate the effectiveness and generalizability of the proposed DRAGON framework, we conduct experiments on four representative COPs across routing, packing and assignment domains:

\noindent$\bullet$ Traveling Salesman Problem (\textbf{TSP}): A classical combinatorial routing problem that seeks the shortest possible tour visiting each city exactly once and returning to the starting point.

\noindent$\bullet$ Capacitated Vehicle Routing Problem (\textbf{CVRP}): Extension of TSP with multiple vehicles and demand constraints. The goal is to minimize total travel distance while satisfying delivery demands.

\noindent$\bullet$ Bin Packing Problem (\textbf{BPP}): Classical combinatorial optimization problem where items of various sizes must be packed into a minimal number of fixed-capacity bins without exceeding their limits.

\noindent$\bullet$ Multiple Knapsack Problem (\textbf{MKP}): A generalization of the 0–1 knapsack problem involving multiple independent knapsacks. Unlike the multi-dimensional knapsack problem~\cite{cacchianiKnapsackProblemsOverview2022}, where each item is constrained by several resource limits, the MKP aims to maximize the total value by optimally assigning items to knapsacks without exceeding their individual capacities.

\subsubsection{Datasets}

Detailed datasets for the aforementioned problems:

\noindent$\bullet$ \textbf{TSP} We use 77 benchmark instances from TSPLIB~\cite{reinelt1991tsplib} of the EUC\_2D type, with sizes ranging from 50 to 20k, covering a broad range of scales and spatial layouts. \textbf{Oracle:} optimal values or lower bounds reported in~\cite{reinelt1991tsplib}.

\noindent$\bullet$ \textbf{CVRP} We select subsets of 19 instances from CVRPLIB, including both X-type~\cite{UCHOA2017845} (up to 1k nodes) and larger XML-type instances~\cite{queiroga2022} distribution, which was synthesized at large scale (up to 5k nodes) by~\cite{zhou2023towards}. \textbf{Oracle:} optimal values or lower bounds reported in~\cite{UCHOA2017845,zhou2023towards}.

\noindent$\bullet$ \textbf{BPP} We adopt instances from FunSearch~\cite{romera2024mathematical} and EoH~\cite{liu2024evolution} of all Weibull-5k, which is reflecting real-world scheduling and allocation scenarios. \textbf{Oracle:} L2 lower bounds determined by~\cite{romera2024mathematical}.

\noindent$\bullet$ \textbf{MKP} Following Google OR-Tools guidelines~\cite{ortools}, we generate 10 synthetic instances by with knapsack capacities uniformly sampled from \( U(100, 500) \), using 10 to 100 knapsacks. Item values and weights drawn uniformly from \(U(1, 100)\). All instances will be released in supplementary JSON files. \textbf{Oracle:} optimal values or uppers bound searched by solver.

\subsubsection{Implementation Details}

We evaluate our framework using four representative LLMs accessed via API: two general-purpose models, OpenAI's \texttt{gpt-4o} and \texttt{gpt-4.1}, and two reasoning-specialized models, OpenAI's \texttt{o3} and DeepSeek's \texttt{r1}. Due to time and cost constraints, we limit our experiments to these models.
Prompting strategies for each stage of DRAGON are illustrated in Figure~\ref{fig:framework}, with complete prompt templates provided in the Appendix (Prompt designs).
We adopt simple global solution initialization algorithms to each problem: 1) Random Insertion~\cite{azar1994lower} for TSP, 2) Greedy Nearest Neighbor heuristic~\cite{mohammed2017solving} for CVRP, and 3) First-Fit Decreasing~\cite{baker1985new} for both BPP and MKP.
For comparative evaluation, we implement OR-Tools~\cite{ortools} solvers as: CVRP is solved via the routing model~\cite{ortools_routing} with \textit{Guided\_Local\_Search} as the metaheuristic, while BPP uses the CP-SAT~\cite{cpsatlp}.
All experiments run on an Ubuntu 20.04 server with an Intel Core i9-10900X (20 cores at 4.6 GHz) without GPU.
For detailed prompt designs used in the decomposition and reconstruction processes, please refer to the Supplementary Materials.

\subsubsection{Feasibility check}

We implement a checker based on the inherent constraints of each COP type. If a solution is found to be infeasible, it is stored in the agent’s memory along with comments explaining the reason for infeasibility. This feedback helps the agent avoid repeating the same mistakes, thereby reducing search time and accelerating convergence to feasible solutions. In the results presented later, we denote an infeasible solution as ``\texttt{infe}'' and use ``\texttt{inf}'' to indicate cases where the solving time exceeds the given time limit $T_{\max}$.

\subsubsection{Metrics}
We evaluate performance using four metrics: (1) Optimality Gap defined as $Gap = \frac{|v - v^*|}{v^*} \times 100\%$, where $v$ and $v^*$ are the objective value and the optimal value, respectively; (2) Running Time (seconds); (3) Input/Output Token Counts; and (4) Number of API Calls. All experiments use consistent settings with a 1-hour time limit $ T_{\max} = 3600 $ and early stopping after $N=5$ consecutive non-improving iterations.

\subsection{Performance Evaluation}

\begin{table}[t]
\centering
\setlength{\tabcolsep}{1mm}
\caption{Average optimality gap (\%) on routing problems using subsets from TSPLIB (EUC\_2D) and CVRPLIB (Set X/XML). }
\begin{tabular}{rccccccc}
\toprule
Method & \multicolumn{5}{c}{TSPLIB (EUC\_2D)} \\\cmidrule(lr){2-7}
(size, k & 0.05-0.5 & 0.5-1  & 1-2 & 2-5 & 5-20 & All \\
\#)      & 42  & 6      & 15    & 7     & 7     & 77 \\
\midrule
OPRO    & \texttt{inf}  & \texttt{inf} & \texttt{inf} & \texttt{inf} & \texttt{inf} & \texttt{inf}  \\
SGE    & \texttt{infe} & \texttt{infe} & \texttt{infe} & \texttt{infe} & \texttt{infe} & \texttt{infe} \\
LMEA &  820.30 & 2279.35 & 3578.21 & \texttt{inf} & \texttt{inf} & \texttt{inf} \\
ReEvo(c)& 12.69 & \textit{16.43} & \textit{17.01} & \texttt{inf} & \texttt{inf} & \texttt{inf} \\
ReEvo(a)& \textbf{8.17} & 16.61 & 18.23 & \textit{21.20} & \textit{18.62} & \textit{13.10} \\
\midrule
DRAGON & \textit{9.74} & \textbf{11.24} & \textbf{15.24} & \textbf{19.37} & \textbf{15.42} & \textbf{12.24} \\
\end{tabular}
\begin{tabular}{rccccccc}
\toprule
Method & \multicolumn{7}{c}{CVRPLIB (X/XML)} \\\cmidrule(lr){2-8} 
(size, k & 0.1-0.2 & 0.2-0.5 & 0.5-1 & 1-2 & 2-5 &  $\ge$5 & All \\
\#)       & 2      & 3      & 5      & 3   & 3  & 3  & 19 \\
\midrule
OR-Tools  & \texttt{inf} & \texttt{inf} & \texttt{inf} & \texttt{inf} & \texttt{inf} & \texttt{inf} & \texttt{inf} \\
ReEvo(a) & \textbf{21.49}  & \textbf{19.27}   & 29.44 & 25.39 & 22.00 & 11.24 & 24.17 \\
\midrule
DRAGON     &  25.56   & 29.35 & \textbf{26.73}  & \textbf{15.51}  & \textbf{15.48} & \textbf{6.45} & \textbf{20.15} \\
\bottomrule
\end{tabular}
\label{tab:result_routing}
\end{table}

To evaluate performance on large-scale routing problems (TSP and CVRP), Table~\ref{tab:result_routing} presents the average optimality gap (\%) across different instance size groups, comparing DRAGON with existing LLM-based solvers, including prompt-based methods (OPRO~\cite{yang2024large}, SGE~\cite{iklassov2024self}) and code-generation approaches (LMEA~\cite{liu2024large}, ReEvo~\cite{ye2024reevo}, with ReEvo(a) using Ant Colony Optimization (ACO) and ReEvo(c) employing a constructive heuristic).
To ensure a fair comparison, all LLM-based methods use \texttt{gpt-4o}, with token usage and inference time recorded. Instances are grouped by node size for conciseness, and detailed results are listed in Supplementary Materials.

OPRO fails to return valid solutions for all groups, as some hard instances in groups exceed the runtime limit, resulting in an infinite gap \texttt{inf}. SGE consistently produces infeasible solutions (\texttt{infe}), likely because it lacks explicit feasibility checks, since its original design is for small problems (under 30 nodes). LMEA handles slightly larger instances but exhibits high gaps and fails beyond 2k-node due to prompt length limits and generation timeouts.

\begin{figure}[h]
    \centering
    \includegraphics[width=0.9\linewidth]{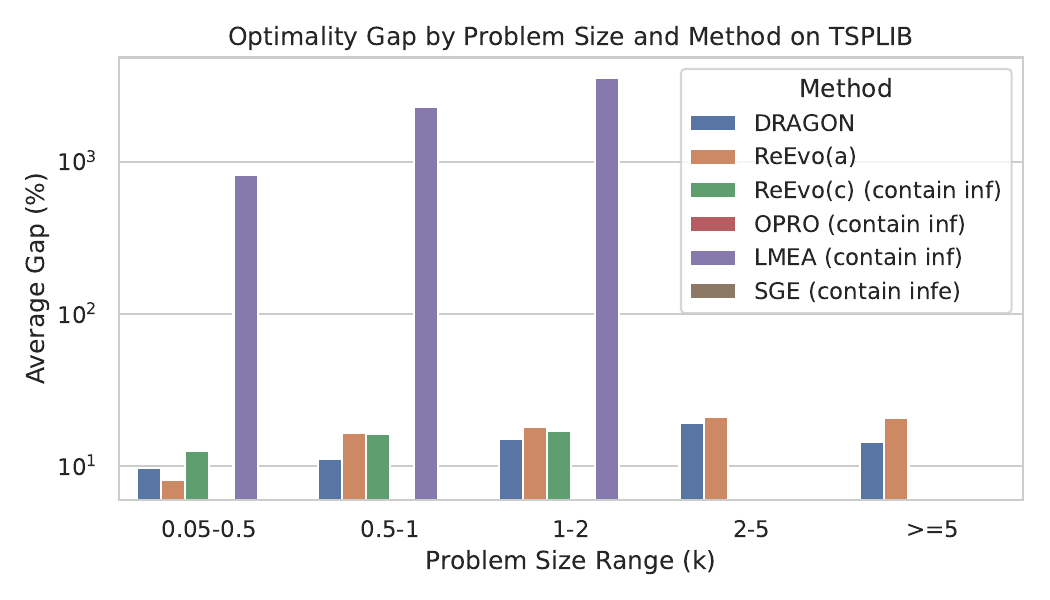}
    \caption{Average Optimality Gap (\%) (log scale) across problem size groups for different methods on TSPLIB.}
    \label{fig:tsp_opt_compare}
\end{figure}

As shown in Figure~\ref{fig:tsp_opt_compare}, the results in TSPLIB reveal that SGE is excluded due to infeasibility, while OPRO is omitted as ``\texttt{inf}'' so that no solution was found due to its slow inference time in all groups. LMEA is able to solve up to medium-sized instances but exhibits relatively large optimality gaps. ReEvo(c) achieves strong results up to the medium-size group, whereas ReEvo(a) excels on smaller instances.
Although DRAGON is not the best on every group, it shows clear advantages on larger-scale problems, driven by its task‑allocation mechanism, where decomposition and reconstruction work effectively together.
\begin{figure}[h]
    \centering
    \includegraphics[width=0.9\linewidth]{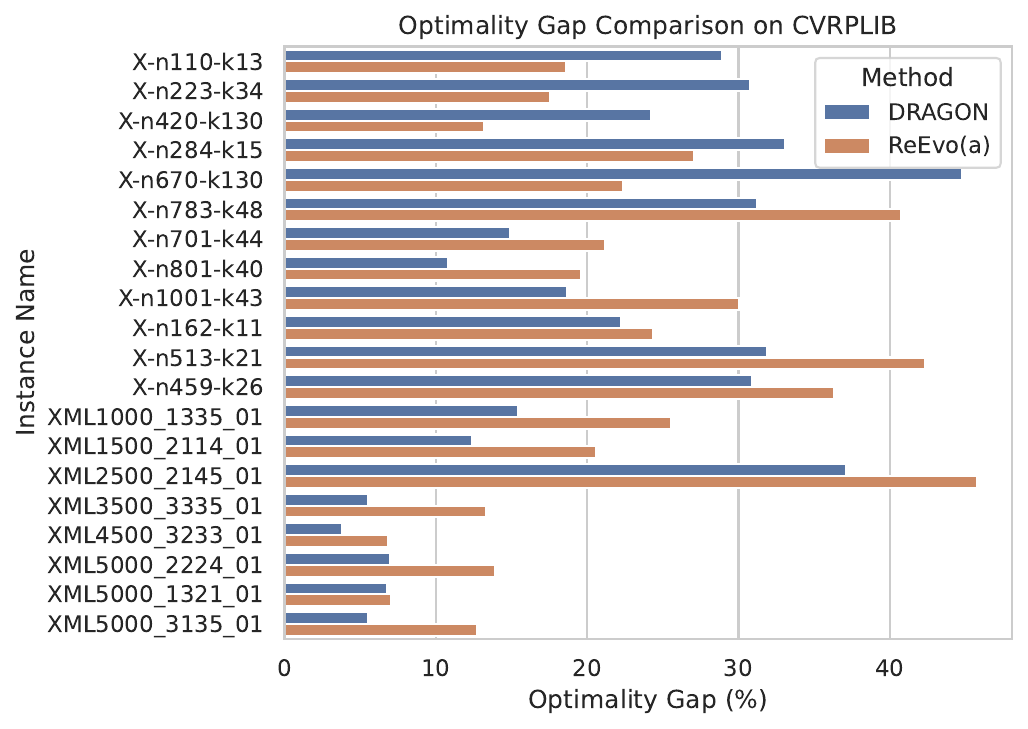}
    \caption{Optimality Gap (\%) for methods on CVRPLIB.}
    \label{fig:cvrp_gap_compare}
\end{figure}
Figure~\ref{fig:cvrp_gap_compare} presents the methods capable of obtaining feasible solutions for CVRP instances. As CVRP is inherently more complex than TSP, DRAGON outperforms ReEvo(a) primarily on large-scale problems, where its decomposition–reconstruction mechanism demonstrates greater advantages.

These findings highlight a major limitation of pure prompt-based methods: their performance degrades significantly as problem size increases. This is mainly due to long context lengths and high decoding cost—making token-by-token generation is unstable when encounter large input loads.
Code-generation methods like ReEvo improve scalability. We tested ReEvo's released heuristic code, including both constructive (ReEvo(c)) and ACO-enhanced (ReEvo(a)) variants. On TSPLIB, DRAGON consistently outperforms ReEvo(c) across all size groups and also surpasses ReEvo(a) on all groups beyond 500, and achieving an average gap reduction of 0.86\%.
On CVRPLIB, where OR-Tools fails to find feasible solutions within time limit, DRAGON leads around 3.98\% lower gaps than ReEvo(a) on all cases. Overall, DRAGON demonstrates the best average performance across routing benchmarks, validating its effectiveness and scalability.

\subsection{Ablation Study}

\subsubsection{Strategies for Decomposition and Reconstruction}

\newcommand{\x}{\ding{55}}
\newcommand{\y}{\ding{51}}

\begin{table}[!t]
\centering
\caption{Domain expertise levels across decomposition–reconstruction strategy combinations. Each cell shows expertise needed: \x\ (none), \y\ (some), \y\y\ (extensive) for decomposition and reconstruction, respectively.}
\begin{tabular}{r|c|c|c}
\toprule
\multirow{2}{*}{\textbf{Decomposition}} & \multicolumn{3}{c}{\textbf{Reconstruction}} \\
\cline{2-4}
 & Heuristic & Solver & LLM \\
\midrule
Random    & (\x, \y ) & (\x , \y\y) & (\x, \x) \\
\hline
Heuristic & (\y, \y) & (\y , \y\y) & (\y , \x) \\
\hline
LLM       & (\x, \y) & (\x , \y\y) & (\x , \x) \\
\bottomrule
\end{tabular}
\label{tab:decomp_reconst_domain_exp}
\end{table}

\begin{figure}[ht]
    \centering
    \includegraphics[width=0.9\linewidth]{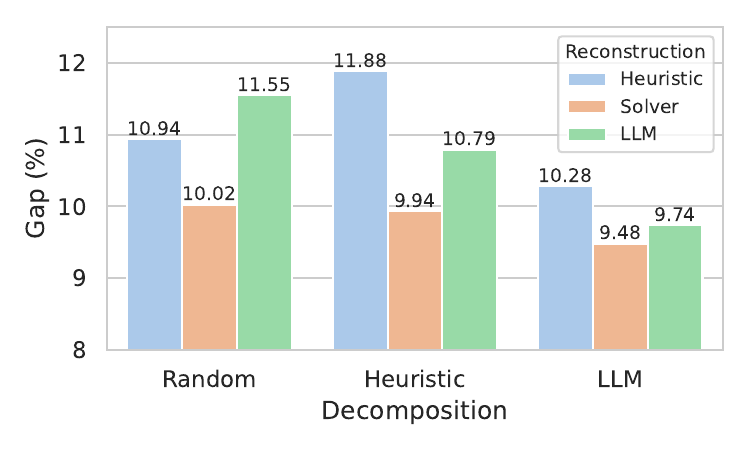}
    \caption{
    Ablation study of DRAGON across 9 decomposition–reconstruction combinations as listed in Table~\ref{tab:decomp_reconst_domain_exp}.
    Bars show the optimality gap (\%) compared to the known optimal value on a TSPLIB subset.
    }
    \label{fig:ablation_strategies}
\end{figure}

DRAGON supports configurable strategies for each task, allowing us to evaluate the impact of different implementations. Table~\ref{tab:decomp_reconst_domain_exp} presents a $3\times3$ combination of decomposition and reconstruction strategies, along with the domain expertise required for each pair, offering insight into their implementation difficulty. 
For decomposition, ``Random'' selects elements arbitrarily, ``Heuristic'' applies domain-specific rules, and ``LLM'' leverages large language models. Similarly for strategies in reconstruction, the only new word ``Solver'' refers to implement general purpose solver.
Figure~\ref{fig:ablation_strategies} shows the performance of DRAGON on a TSPLIB subset using these nine combinations.

Among reconstruction strategies, the solver-based approach achieves the best performance by optimally solving small size local COPs. However, this method has key drawbacks: (1) in routing problems for example, enforcing must-visit edges via zero distances with penalty trick may not lead to the constraint satisfied  outcomes, and (2) encoding custom constraints across diverse COPs requires substantial domain expertise, as reflected in Table~\ref{tab:decomp_reconst_domain_exp}.
Despite requiring the most domain knowledge, the ``Heuristic$\times$Solver'' combination does not yield the best results.
In contrast, our DRAGON pipeline ``LLM$\times$LLM'' outperforms it, highlighting the effectiveness of LLM-based decomposition agent in capturing promising substructures.
Notably, reconstruction agent offers ease of implementation and integrates naturally with  decomposition agent. Hence, DRAGON strikes a strong balance between performance and usability.

\subsubsection{Impact of LLM backbones}

\begin{figure}[t]
    \centering
    \includegraphics[width=0.95\linewidth]{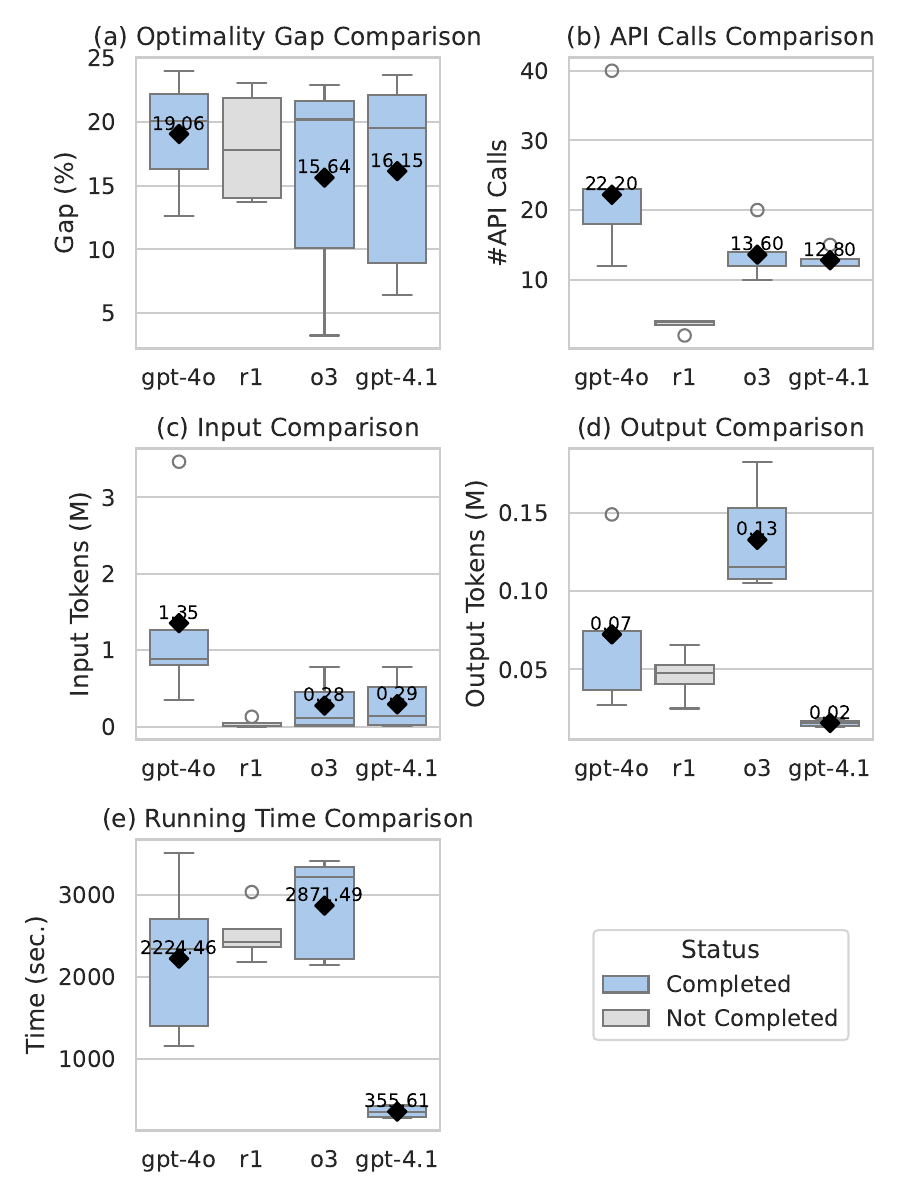}
    \caption{Comparison of LLM models by: (a) optimality gap, (b) number of API calls, (c) input tokens, (d) output tokens, (e) running time. We evaluate general-purpose models (\texttt{gpt-4o}, \texttt{gpt-4.1}) and reasoning-oriented models (\texttt{o3}, \texttt{r1}). Gray bars indicate incomplete results due to token limit violations.}
    \label{fig:llm_backbones}
\end{figure}

As shown in Figure~\ref{fig:llm_backbones}, we evaluate the impact of different LLM backbones within the DRAGON framework, including general-purpose models (OpenAI \texttt{gpt-4o}, \texttt{gpt-4.1}) and reasoning models (OpenAI \texttt{o3}, DeepSeek \texttt{r1}). 
Due to cost considerations, we do not evaluate all TSPLIB instances. As \texttt{r1} exceeds the model's input token limit of 65,536 on instances larger than 10k nodes, it fails to produce responses in the largest case (listed in Appendix (Experiment results)).

While \texttt{o3} achieves the lowest average optimality gap, it incurs extremely high output token counts. Given that OpenAI's API typically prices output tokens at 4 times the rate of input tokens, this leads to significantly higher costs and longer inference time. 
\texttt{r1} exhibits a relatively low number of API calls, suggesting that each call takes longer—likely due to more intensive reasoning per iteration.
Considering the trade-off between objective quality, inference speed, and token efficiency, \texttt{gpt-4.1} achieves the best overall balance and emerges as the current most practical choice.

\subsection{Generalization Case Study}

To assess the robustness of DRAGON, we evaluate its performance across more domains, it highlights the method’s ability to generalization for the subsequent results.

\subsubsection{Bin Packing Problem}

\begin{table}[!t]
\centering
\caption{Average optimality gap (\%) w.r.t. L2 lower bound and inference time (sec.) on the Weibull-5K dataset.}
\setlength{\tabcolsep}{0.9mm}
\begin{tabular}{rccccc}
\toprule
Method &ReEvo(a) &FunSearch &EoH &EoH expert &Ours \\\midrule
Gap (\%) $\downarrow$ &3.46 &0.69 &0.66 &0.55 &\textbf{0.33} \\
Time (sec.) $\downarrow$ & 56.677 & 2.292 & - & - & 487.873 \\
\bottomrule
\end{tabular}
\label{tab:result_bpp}
\end{table}

In addition to routing problems, we evaluate DRAGON on the packing domain. As shown in Table~\ref{tab:result_bpp}, DRAGON consistently outperforms ReEvo(a), FunSearch, EoH, and EoH expert on the Weibull-5K dataset. We omit OR-Tools solvers (both MP
and CP-SAT backends) as they failed to produce feasible solutions before timeouts.
 
Compared to routing, packing problems are generally easier for LLMs to reason about due to their more intuitive and straightforward constraints. For example, merging static segments ($\mathbf{s}$) into packed items requires less complex reasoning, allowing DRAGON to scale more effectively to larger instances. Results for EoH and EoH expert are reported directly from~\cite{liu2024evolution}, where running time was not provided.
While DRAGON achieves the lowest optimality gap, it is slower than code-generation based methods. However, code-generation approaches come with high API costs and substantial time and resources to evolve or fine-tune high-quality heuristic functions. These overheads are not captured in current comparison.
Each paradigm has its strengths. For large batches of similar instances, code-generation methods are effective as the learned heuristic can be reused across the dataset. In contrast, prompt-based approaches like DRAGON are better suited for dynamic environments where instance structures vary, offering flexibility without the need for retraining or code evolution.

\subsubsection{Multiple Knapsack Problem}

\begin{figure}[ht]
    \centering
    \includegraphics[width=0.9\linewidth]{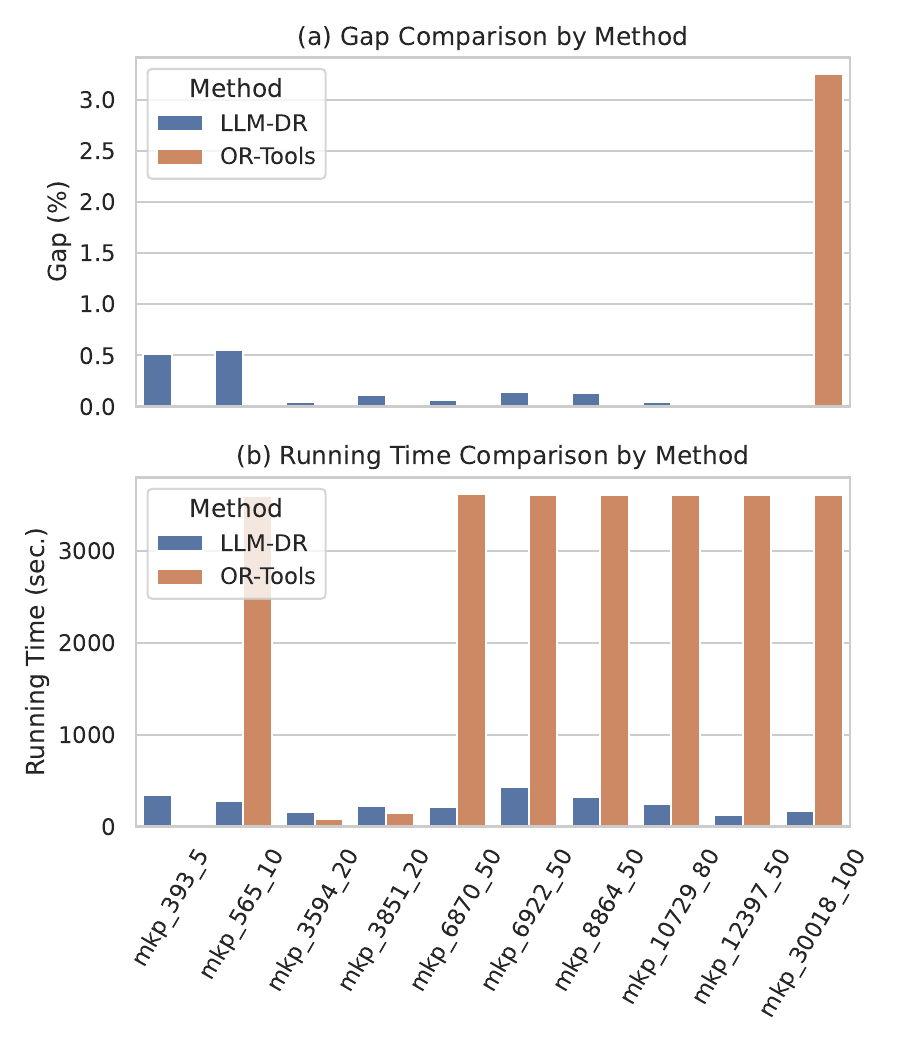}
    \caption{Comparison on synthetic Multiple Knapsack Problem instances between DRAGON and OR-Tools (CP-SAT).}
    \label{fig:mkp_compare}

\end{figure}

We compare DRAGON with the CP-SAT on a series of synthetic MKP instances. Each instance is named using the format \texttt{mkp\_$n_{item}$\_$n_{knapsack}$}, where the decision space grows as $n_{item} \times n_{knapsack}$, hence the maximum of decison size here is around 3M.

As shown in Figure~\ref{fig:mkp_compare}, CP-SAT typically achieves lower optimality gaps and can even find optimal solutions when the instance size is manageable. However, its performance starts to degrade on larger instances, evident from the largest case, where the gap spikes to around 3\% and it hits the time limit.
DRAGON performs favorably across all instances. While its gap is generally slightly higher than CP-SAT on small cases, it remains consistently below 0.5\%, even on large-scale instances.
In terms of trade-offs between performance and efficiency, DRAGON offers strong scalability and competitive quality, delivering high-quality solutions within reasonable reasoning time, making it highly effective for large-scale MKPs where traditional solvers may struggle.

\subsection{Limitations and Future Directions}

While DRAGON demonstrates strong potential in solving large-scale COPs, several limitations remain to be addressed.
The quality of the reconstructed global solution is sensitive to both the decomposition strategy and the prompt formulation. Inappropriate segmentation or poorly structured prompts may lead to suboptimal or degenerated solutions, emphasizing the need for more robust and adaptive decomposition mechanisms.
Moreover, the iterative nature of DRAGON introduces substantial computational overhead due to the large context required during each divide-and-conquer cycle, resulting in extended runtime. Although the current pipeline may not be efficient for small-scale problems, it proves particularly advantageous for super large-scale scenarios where traditional solvers and other learning-based methods struggle to deliver feasible or high-quality solutions across all testing tasks.

DRAGON also demonstrates that purely language-based agents can effectively address large-scale COPs, outperforming other prompt-based or code-generation-based approaches. Additionally, incorporating external optimization tools such as OR-Tools within the reconstruction stage enhances local refinement. However, the use of such tools often demands additional modeling expertise, especially for non-standard constrained sub-problem settings.

In the on-going work, we are integrating more external tools to automatically model customized constraints not only within the reconstruction agent but also into the decomposition process, with the goal of further reducing runtime and API costs while maintaining or improving solution quality. We also envision introducing a central coordination agent to better manage the interaction among decomposition and reconstruction agents, thereby improving the overall consistency and efficiency of the optimization process.
\section{Conclusion}

This paper presents DRAGON, a novel framework that leverages LLM agents to solve large-scale COPs via an iterative decomposition reconstruction process inspired by divide-and-conquer. Experiments across diverse COP domains (e.g., routing and packing) show that DRAGON consistently achieves strong performance, especially on extreme large instances where traditional solvers or basic prompt-based methods often fail due to scalability or timeouts.
By decomposing complex problems into context manageable subproblems, DRAGON enables LLMs to reason effectively over intricate structures, identifying promising regions, and reconstructing feasible global solutions. This training-free, modular, and solver-agnostic approach supports flexible constraint handling and adapts easily to diverse COPs without specialized algorithm design.
Although not optimized for speed, DRAGON avoids expensive code search and offers a strong balance between solution quality, adaptability, and cost, making it particularly well-suited for dynamic and large-scale settings. 
Future work will focus on enhancing decomposition strategies, improving prompt robustness, and integrating hybrid solvers to explore promising regions more efficiently. 
Overall, DRAGON establishes a solid foundation toward general-purpose, interpretable, and scalable LLM-based optimization.



\begin{acks}
This research/project is supported by the National Research Foundation, Singapore under its AI Singapore Programme (AISG Award No: AISG3-RPGV-2025-017, AISG3-RP-2025-036-USNSF).
For Supplementary Materials: \url{https://arxiv.org/abs/2601.06502} 
\end{acks}



\bibliographystyle{ACM-Reference-Format} 
\bibliography{aamas}

\appendix

\renewcommand{\thefigure}{A\arabic{figure}}
\renewcommand{\thetable}{A\arabic{table}}
\setcounter{figure}{0}
\setcounter{table}{0}

\section{Prompt Designs}

\tcbset{
  boxstyle1/.style={
    colback=blue!5,
    colframe=blue!70!black,
    arc=6pt,
    boxrule=0.5pt,
    fontupper=\ttfamily,
    enhanced,
    breakable,
    left=10pt, right=10pt, top=10pt, bottom=10pt,
    width=\linewidth,
  },
  boxstyle2/.style={
    colback=orange!5,
    colframe=orange!50!black,
    arc=6pt,
    boxrule=0.5pt,
    fontupper=\ttfamily,
    enhanced,
    breakable,
    left=10pt, right=10pt, top=10pt, bottom=10pt,
    width=\linewidth,
  },
    boxstyle3/.style={
    colback=green!5,
    colframe=green!50!black,
    arc=6pt,
    boxrule=0.5pt,
    fontupper=\ttfamily,
    enhanced,
    breakable,
    left=10pt, right=10pt, top=10pt, bottom=10pt,
    width=\linewidth,
  },
  boxstyle4/.style={
    colback=yellow!5,
    colframe=yellow!50!black,
    arc=6pt,
    boxrule=0.5pt,
    fontupper=\ttfamily,
    enhanced,
    breakable,
    left=10pt, right=10pt, top=10pt, bottom=10pt,
    width=\linewidth,
  }
}

\subsection{General DRAGON Prompts}

To implement DRAGON, we detail the design of the prompts used by the framework, which consists of two main agents: the \textbf{Decomposer} and the \textbf{Reconstructor}. These LLM prompts are designed to be general-purpose and adaptable across various combinatorial optimization problems (COPs). They contain placeholders that should be customized based on the specific COP being addressed.

\subsubsection{Decomposer's System Prompt}
~
\begin{tcolorbox}[boxstyle1, title=Decomposer System Prompt]
You are an expert in optimization with smart heuristics. The user will provide you an initial solution, whose data are formatted as \{\textcolor{brown}{METADATA\_FORMAT}\}. Help the user analysis the initial solution and point out which part is yet to be optimized using any creative heuristic methods you can. Ensure to output the answer in the specified required format.

~

The problem inputs are:

\{\textcolor{brown}{INPUT\_DATA}\}

~

**Format**: Return only the node index, enclosed in <sub> and </sub>, separated by commas.

For example: <sub>1,2,3</sub>

\end{tcolorbox}

The placeholders above should be filled as follows:
\begin{itemize}
    \item \texttt{METADATA\_FORMAT}: A list of key-value pairs describing the structure of the solution input.
    \item \texttt{INPUT\_DATA}: The actual raw input data to be passed to the LLM.
\end{itemize}

To align with our implementation, we recommend structuring the input data in a standardized JSON format, as illustrated in Table~\ref{tab:dataformat}.
\begin{table}[htbp]
    \setlength{\tabcolsep}{10.5pt} 
    \centering
    \caption{Required keys in JSON format for different COPs. \y indicates mandatory fields, while \textcolor{lightgray}{\y} marks optional fields that are applicable in routing problems with additional constraints.}
    \begin{tabular}{r|cccc}
    \toprule
    Key & TSP & CVRP & BPP & MKP \\
    \midrule
    ``name'' & \y & \y & \y & \y \\
    ``type'' & \y & \y & \y & \y \\
    ``num'' & \y & \y & \y & \y \\
    ``depot'' &  & \y &  &  \\
    ``x'' & \y & \y &  &  \\
    ``y'' & \y & \y &  &  \\
    ``weights'' &  &  & \y & \y \\
    ``values'' &  &  &  & \y \\
    ``capacity'' &  & \y & \y & \y \\
    ``demand'' &  & \y &  &  \\
    ``link'' & \textcolor{lightgray}{\y} & \textcolor{lightgray}{\y} \\
    \bottomrule
    \end{tabular}
    \label{tab:dataformat}
\end{table}
Value types in the JSON schema are:
\begin{itemize}
    \item \textbf{String} (e.g., ``name'', ``type''),
    \item \textbf{Integer} (e.g., ``depot'' for CVRP, ``capacity'' for BPP, ``num'' refer to node size for TSP/CVRP and item size for BPP/MKP),
    \item \textbf{List of integers} (e.g., ``x'', ``y'' for routing, ``demand'', ``capacity'' for MKP).
\end{itemize}
For example of CVRP metadata:
\begin{tcolorbox}[boxstyle3, title=Example JSON for CVRP]
\begin{verbatim}
{
  "name": "CVRP-Example-001",
  "type": "cvrp",
  "num": 6,
  "depot": 0,
  "x": [50, 20, 40, 60, 80, 30],
  "y": [50, 70, 60, 40, 30, 90],
  "capacity": 100,
  "demand": [0, 10, 20, 30, 25, 15],
  "link": [
       [0, 5], 
       [4, 2]
      ]
}
\end{verbatim}
\end{tcolorbox}

\textbf{Decomposer's User Prompt}:

\begin{tcolorbox}[boxstyle2, title=Decomposer User Prompt]
For the given optimization problem (\{\textcolor{brown}{PROBLEM\_TYPE}\}), my current solution is:

~

\{\textcolor{brown}{SOLUTION}\}

~

These \{\textcolor{brown}{SOLUTION\_MARK}\} are not optimal. I want to improve them by remove some \{\textcolor{brown}{ELEMENT}\} from the current solution, reconstruct them optimally.

Please identify no more than \{\textcolor{brown}{NUM}\} \{\textcolor{brown}{ELEMENT}\} that could be improved.

\end{tcolorbox}

Here’s how to define the placeholders:
\begin{itemize}
    \item \texttt{PROBLEM\_TYPE}: Set as ``tsp'', ``cvrp'', ``bpp'', or ``mkp''.
    \item \texttt{SOLUTION}: The current solution in a structured format (e.g., XML or JSON). For CVRP, the XML looks like:
\end{itemize}

\begin{tcolorbox}[boxstyle3, title=Example XML for CVRP]
<sol>

~<route>0,2,3,0</route>

~<route>0,1,5,4,0</route>

</sol>
\end{tcolorbox}

\begin{table}[htbp]
    \centering
    \caption{Mapping of user prompt placeholders for COPs.}
    \begin{tabular}{r|cccc}
    \toprule
    Placeholder & TSP & CVRP & BPP & MKP \\
    \midrule
    \texttt{SOLUTION\_MARK} & ``route'' & ``route'' & ``pack'' & ``pack''  \\
    \texttt{ELEMENT} & ``node'' & ``node'' & ``item'' & ``item''  \\
    \bottomrule
    \end{tabular}
    \label{tab:user_args}
\end{table}

We set \texttt{NUM} = 20 in all experiments. 
We recommend not exceeding 50, as the identified elements will be passed to the reconstructor, which may face difficulty handling larger inputs.

\subsubsection{Reconstructor's System Prompt}

\begin{tcolorbox}[boxstyle1, title=Reconstructor System Prompt]
Act as an expert in combinatorial optimization. Your goal is find the best solution with given constraints, use any heuristic as you can.

~

**Goal: Find the best solution, avoid the infeasible solutions.**

**Requirements:**

- \{\textcolor{brown}{PROBLEM\_SPECIFIC\_REQUIREMENTS}\}

- Check the feasibility of your solution to ensure the above conditions.

~

OUTPUT FORMAT: After your solving, present your answer as:

~

\{\textcolor{brown}{SOLUTION\_FORMAT}\}

\end{tcolorbox}

Where the \texttt{PROBLEM\_SPECIFIC\_REQUIREMENTS} are:

\begin{itemize}
    \item TSP:
    \begin{itemize}
        \item Must visit all the points exactly once, except the depot.
        \item The start and end points must remain fixed.
        \item The fixed path prevents any other visits in between; allow reversing of the fixed path where necessary.
    \end{itemize}
    \item CVRP:
    \begin{itemize}
        \item Must visit all the customer nodes exactly once.
        \item Each route must start and end at the depot.
        \item Vehicle capacity must not be exceeded.
        \item The fixed path prevents any other visits in between; allow reversing of the fixed path where necessary.
    \end{itemize}
    \item BPP:
    \begin{itemize}
        \item Packed items' total weight in each bin must be less than or equal to the bin's capacity.
    \end{itemize}
    \item MKP:
    \begin{itemize}
        \item Selected items' total weight must be less than or equal to the knapsack's capacity.
    \end{itemize}
\end{itemize}
And the \texttt{SOLUTION\_FORMAT} should follow:
\begin{itemize}
    \item TSP:
    \begin{tcolorbox}[boxstyle4]
        <sol>

        ~<route>0,1,2,...,0</route>

        </sol>
    \end{tcolorbox}

    \item CVRP:
    \begin{tcolorbox}[boxstyle4]
        <sol>

        ~<route>0,1,2,...,0</route>

        ~<route>0,3,4,...,0</route>

        ~...

        </sol>
    \end{tcolorbox}

    \item BPP:
    \begin{tcolorbox}[boxstyle4]
        <sol>

        ~<bin\_0>0,1,2,...</bin\_0>

        ~<bin\_1>3,4,...</bin\_1>

        ~...

        </sol>

~

        Where bin\_i is the i-th bin.
    \end{tcolorbox}

    \item MKP:
    \begin{tcolorbox}[boxstyle4]
        <sol>

        ~<knapsack\_0>0,1,2,...</knapsack\_0>

        ~<knapsack\_1>3,4,...</knapsack\_1>

        ~...

        </sol>

~

        Where knapsack\_i is the i-th knapsack.
    \end{tcolorbox}
\end{itemize}

\textbf{Reconstructor's User Prompt}:

\begin{tcolorbox}[boxstyle2, title=Reconstructor User Prompt]
Given a set of \{\textcolor{brown}{ELEMENTS}\}, your task is to find the \{\textcolor{brown}{OPTIMIZATION\_OBJECT}\} while respecting the following constraints.

~

\{\textcolor{brown}{CONSTRAINTS}\}

~

Current solution:

\{\textcolor{brown}{SOLUTION}\}

\end{tcolorbox}
Where \texttt{ELEMENT} is set as:
\begin{itemize}
    \item ``nodes'' for TSP and CVRP
    \item ``items'' for BPP and MKP
\end{itemize}
The \texttt{OPTIMIZATION\_OBJECT} is set as:
\begin{itemize}
    \item TSP: ``Find the shortest possible tour that visits all nodes exactly once and returns to the starting depot.''
    \item CVRP: ``Design a set of routes to deliver all customer demands using vehicles with limited capacity, minimizing the total distance while visiting each customer exactly once.''
    \item BPP: ``Given a set of items with their weights, your task is to find the best packing solution that minimizes the number of bins used while respecting each bin's capacity.''
    \item MKP: ``Given a set of items with their values and weights, your task is to find the best packing solution that maximizes the total value while respecting each knapsack's capacity.''
\end{itemize}
The \texttt{CONSTRAINTS} are:
\begin{itemize}
    \item Routing problem (TSP and CVRP):
    \begin{itemize}
        \item ``Fixed visiting path as (1,3), (5,9), ...'' (i.e., some segments in the tour must be preserved).
    \end{itemize}
    \item Packing problem (BPP and MKP):
    \begin{itemize}
        \item Sum all items in the static segment $\mathbf{s}$ of each bin or knapsack, and treat them as a new bulky item, and set its weight as the sum of item weights. For MKP, further set the bulky item's weight and value as the respective sums.
    \end{itemize}
\end{itemize}
\texttt{SOLUTION} in positive experience storage should include the current solution in the same XML format as required by the \texttt{SOLUTION\_FORMAT} in the system prompt.
If you have collected infeasible solutions by previous rounds, you can continue to append your user prompt:
\begin{tcolorbox}[boxstyle2, title=Reconstructor User Prompt (append revision)]

...

Analysis the following infeasible solution(s) and generate a new solution to meet the given constraints.

~

\{\textcolor{brown}{INFEASIBLE\_SOL1}\}

~

\{\textcolor{brown}{INFEASIBLE\_SOL2}\}

~

\{\textcolor{brown}{INFEASIBLE\_SOL3}\}

...
\end{tcolorbox}
Where the template for negative experience \texttt{INFEASIBLE\_SOLx} can be:
\begin{tcolorbox}[boxstyle4]
    <sol>
    
    ...
    
    </sol>

    <reason>

        Missing visit node(s): 3, 7 ....

    </reason>
\end{tcolorbox}

\section{Detailed Experimental Results}

\subsection{TSPLIB Results}
We then present full results on benchmarking instances from TSPLIB in Tables~\ref{tab:tsp_llmda}–\ref{tab:tsp_reevoa}.
OPRO fails to produce valid solutions across all groups, as several hard instances exceed the runtime limit, leading to infinite gaps (inf). SGE consistently returns infeasible solutions (infe), likely due to the absence of explicit feasibility checks, as it was originally designed for small-scale problems (fewer than 30 nodes). LMEA can handle moderately larger instances, but suffers from high optimality gaps and fails on instances beyond 2k nodes, primarily due to prompt length constraints and generation timeouts.

\subsection{CVRPLIB Results}
Full results of CVRPLIB instances are listed in Tables~\ref{tab:cvrp_llmda}–\ref{tab:cvrp_reevo}.

\subsection{Ablation Studies}

Ablation results analyzing decomposition and reconstruction strategies are provided in Table~\ref{tab:ablation_strategy}, and ablation results of LLM models are provided in Table~\ref{tab:ablation_llm}.

\subsection{Generalization Case Studies}
\textbf{BPP:} Results are listed in Tables~\ref{tab:bpp_llmdr}-\ref{tab:bpp_reevo}.
\textbf{MKP:} Performance are shown in Tables~\ref{tab:mkp_llmdr}-\ref{tab:mkp_solver}.



\begin{table*}[h]
\renewcommand{\arraystretch}{0.55}
\centering
\small
\caption{Result of \textbf{DRAGON} on the TSPLIB \texttt{EUC\_2D} subset.}
\label{tab:tsp_llmda}
\begin{tabular}{crrrrrrrr}
\toprule
Size (k) & Name & Lower bound & Objective value & Gap (\%) & \#API calls & Input tokens (k) & Output tokens (k) & Running time (sec.) \\
\midrule
\multirow{6}{*}{$<$0.1}	& berlin52 & 7542 & 8773 & 16.322 & 18 & 352.398 & 27.373 & 1153.372 \\
 & eil51 & 426 & 440 & 3.286 & 25 & 355.136 & 33.297 & 1775.789 \\
 & eil76 & 538 & 598 & 11.152 & 27 & 584.812 & 45.871 & 1678.698 \\
 & pr76 & 108159 & 114421 & 5.790 & 16 & 293.463 & 20.841 & 929.784 \\
 & rat99 & 1211 & 1402 & 15.772 & 18 & 257.759 & 26.649 & 1827.705 \\
 & st70 & 675 & 757 & 12.148 & 24 & 409.572 & 35.584 & 1757.166 \\
\midrule
\multirow{21}{*}{0.1-0.2}	& bier127 & 118282 & 135482 & 14.542 & 29 & 330.866 & 34.255 & 1957.364 \\
 & ch130 & 6110 & 6503 & 6.432 & 32 & 533.508 & 60.685 & 2831.500 \\
 & ch150 & 6528 & 7185 & 10.064 & 37 & 585.292 & 56.462 & 2719.040 \\
 & d198 & 15780 & 17621 & 11.667 & 46 & 871.951 & 67.258 & 3499.006 \\
 & eil101 & 629 & 694 & 10.334 & 40 & 705.357 & 70.520 & 2621.478 \\
 & kroA100 & 21282 & 23160 & 8.824 & 40 & 690.147 & 59.315 & 2354.898 \\
 & kroA150 & 26524 & 28278 & 6.613 & 36 & 502.292 & 47.662 & 1928.851 \\
 & kroB100 & 22141 & 23417 & 5.763 & 40 & 507.247 & 64.044 & 3507.437 \\
 & kroB150 & 26130 & 29267 & 12.005 & 9 & 223.708 & 17.918 & 613.975 \\
 & kroC100 & 20749 & 22964 & 10.675 & 20 & 370.270 & 32.169 & 1566.927 \\
 & kroD100 & 21294 & 22843 & 7.274 & 28 & 433.136 & 40.538 & 2048.876 \\
 & kroE100 & 22068 & 23402 & 6.045 & 23 & 415.897 & 37.632 & 1696.224 \\
 & lin105 & 14379 & 15979 & 11.127 & 17 & 198.105 & 22.608 & 1115.946 \\
 & pr107 & 44303 & 44925 & 1.404 & 22 & 200.629 & 27.529 & 1556.138 \\
 & pr124 & 59030 & 62151 & 5.287 & 22 & 386.629 & 35.707 & 1380.136 \\
 & pr136 & 96772 & 101634 & 5.024 & 18 & 301.971 & 24.507 & 1008.675 \\
 & pr144 & 58537 & 62220 & 6.292 & 18 & 347.802 & 25.506 & 932.202 \\
 & pr152 & 73682 & 77233 & 4.819 & 13 & 202.484 & 14.085 & 745.549 \\
 & rat195 & 2323 & 2588 & 11.408 & 26 & 438.158 & 47.778 & 2309.684 \\
 & rd100 & 7910 & 8865 & 12.073 & 27 & 555.131 & 48.948 & 1984.560 \\
 & u159 & 42080 & 47756 & 13.489 & 5 & 95.530 & 6.773 & 451.494 \\
\midrule
\multirow{15}{*}{0.2-0.5} & a280 & 2579 & 2901 & 12.485 & 19 & 358.893 & 27.183 & 988.452 \\
 & d493 & 35002 & 39490 & 12.822 & 38 & 685.977 & 67.881 & 3444.888 \\
 & fl417 & 11861 & 12776 & 7.714 & 22 & 350.584 & 28.860 & 1451.702 \\
 & gil262 & 2378 & 2644 & 11.186 & 9 & 223.568 & 19.183 & 836.437 \\
 & kroA200 & 29368 & 31920 & 8.690 & 31 & 603.447 & 45.735 & 2069.374 \\
 & kroB200 & 29437 & 33157 & 12.637 & 40 & 803.297 & 74.733 & 3517.805 \\
 & lin318 & 42029 & 46716 & 11.152 & 31 & 660.784 & 46.946 & 2264.698 \\
 & pcb442 & 50778 & 58232 & 14.680 & 35 & 760.501 & 60.740 & 2542.813 \\
 & pr226 & 80369 & 81657 & 1.603 & 19 & 336.121 & 23.377 & 807.854 \\
 & pr264 & 49135 & 52991 & 7.848 & 29 & 574.093 & 49.003 & 1840.055 \\
 & pr299 & 48191 & 53787 & 11.612 & 18 & 333.709 & 30.133 & 997.101 \\
 & pr439 & 107217 & 121879 & 13.675 & 18 & 364.550 & 36.918 & 1166.819 \\
 & rd400 & 15281 & 17215 & 12.656 & 13 & 261.578 & 20.461 & 1039.604 \\
 & ts225 & 126643 & 144636 & 14.208 & 26 & 517.690 & 46.750 & 2419.392 \\
 & tsp225 & 3919 & 4324 & 10.334 & 9 & 152.992 & 15.664 & 789.163 \\
\midrule
\multirow{6}{*}{0.5-1} & d657 & 48912 & 54577 & 11.582 & 26 & 618.878 & 54.094 & 1999.693 \\
 & p654 & 34643 & 37476 & 8.178 & 19 & 518.997 & 32.475 & 1282.118 \\
 & rat575 & 6773 & 7590 & 12.063 & 49 & 1085.435 & 82.549 & 3511.544 \\
 & rat783 & 8806 & 9968 & 13.196 & 44 & 956.776 & 95.982 & 3505.717 \\
 & u574 & 36905 & 40922 & 10.885 & 28 & 779.953 & 40.176 & 1516.220 \\
 & u724 & 41910 & 46740 & 11.525 & 16 & 342.380 & 18.580 & 982.001 \\
\midrule
\multirow{15}{*}{1-2} & d1291 & 50801 & 60404 & 18.903 & 40 & 1640.458 & 61.223 & 2530.448 \\
 & d1655 & 62128 & 73834 & 18.842 & 14 & 456.675 & 32.196 & 1164.084 \\
 & fl1400 & 20127 & 21521 & 6.926 & 33 & 920.564 & 98.208 & 3246.920 \\
 & fl1577 & 22249 & 25526 & 14.729 & 5 & 69.703 & 10.569 & 243.459 \\
 & nrw1379 & 56638 & 63780 & 12.610 & 13 & 671.878 & 32.245 & 1241.157 \\
 & pcb1173 & 56892 & 65849 & 15.744 & 9 & 462.186 & 18.142 & 625.697 \\
 & pr1002 & 259045 & 293971 & 13.483 & 41 & 1609.575 & 62.281 & 2218.305 \\
 & rl1304 & 252948 & 302389 & 19.546 & 33 & 895.836 & 69.866 & 2343.211 \\
 & rl1323 & 270199 & 325813 & 20.583 & 22 & 846.612 & 42.888 & 1383.468 \\
 & rl1889 & 316536 & 376474 & 18.936 & 29 & 867.053 & 98.208 & 2627.414 \\
 & u1060 & 224094 & 248657 & 10.961 & 16 & 349.302 & 33.205 & 986.517 \\
 & u1432 & 152970 & 172925 & 13.045 & 22 & 1033.374 & 39.479 & 1447.544 \\
 & u1817 & 57201 & 68175 & 19.185 & 18 & 798.746 & 42.234 & 1986.882 \\
 & vm1084 & 239297 & 267949 & 11.973 & 2 & 58.211 & 6.195 & 169.439 \\
 & vm1748 & 336556 & 380867 & 13.166 & 43 & 1271.237 & 88.938 & 3346.815 \\
\midrule
\multirow{7}{*}{2-5} & d2103 & 80450 & 99775 & 24.021 & 23 & 888.182 & 74.305 & 2711.948 \\
 & fl3795 & 28772 & 34268 & 19.102 & 26 & 1275.895 & 89.607 & 2589.477 \\
 & fnl4461 & 152566 & 207168 & 35.789 & 18 & 1289.366 & 34.595 & 1760.015 \\
 & pcb3038 & 137694 & 158664 & 15.229 & 18 & 714.400 & 67.205 & 2618.376 \\
 & pr2392 & 378032 & 430293 & 13.824 & 38 & 1663.085 & 69.731 & 3113.234 \\
 & u2152 & 64253 & 77075 & 19.955 & 33 & 1061.294 & 108.541 & 3208.794 \\
 & u2319 & 234256 & 252285 & 7.696 & 21 & 632.640 & 78.352 & 2206.402 \\
\midrule
\multirow{2}{*}{5-10} & rl5915 & 655530 & 688752 & 5.068 & 14 & 1385.583 & 67.902 & 1733.749 \\
 & rl5934 & 556045 & 679642 & 22.228 & 18 & 3465.698 & 149.070 & 2340.457 \\
\midrule
\multirow{5}{*}{$\ge$10} & brd14051 & 469445 & 530915 & 13.094 & 24 & 2502.501 & 37.871 & 1950.459 \\
 & d15112 & 1573152 & 1780905 & 13.206 & 26 & 3877.022 & 49.884 & 2741.064 \\
 & d18512 & 645488 & 733860 & 13.691 & 2 & 525.051 & 2.558 & 1012.969 \\
 & rl11849 & 923368 & 1108759 & 20.078 & 12 & 1261.444 & 36.615 & 1398.704 \\
 & usa13509 & 19982889 & 22827898 & 14.237 & 22 & 3515.226 & 117.270 & 2748.211 \\
\bottomrule
\end{tabular}
\end{table*}

\begin{table*}[h]
\centering
\renewcommand{\arraystretch}{0.55}
\centering
\small
\caption{Result of LMEA on the TSPLIB \texttt{EUC\_2D} subset.}
\label{tab:tsp_lmea}
\begin{tabular}{crrrrrrrr}
\toprule
Size (k) & Name & Lower bound & Objective value & Gap (\%) & \#API calls & Input tokens (k) & Output tokens (k) & Running time (sec.) \\
\midrule
\multirow{6}{*}{$<$0.1} & berlin52 & 7542 & 22985 & 204.760 & 16 & 48.560 & 109.932 & 1633.713 \\
 & eil51 & 426 & 1194 & 180.282 & 24 & 71.568 & 174.415 & 1953.738 \\
 & eil76 & 538 & 1828 & 239.777 & 29 & 114.028 & 272.653 & 2818.638 \\
 & pr76 & 108159 & 396874 & 266.936 & 37 & 150.923 & 346.013 & 3486.687 \\
 & rat99 & 1211 & 6349 & 424.277 & 19 & 91.314 & 237.203 & 3511.408 \\
 & st70 & 675 & 2710 & 301.481 & 33 & 122.232 & 289.233 & 3575.126 \\
\midrule
\multirow{21}{*}{0.1-0.2} & bier127 & 118282 & 522377 & 341.637 & 23 & 140.829 & 306.311 & 3585.385 \\
 & ch130 & 6110 & 38924 & 537.054 & 22 & 131.648 & 226.152 & 3339.959 \\
 & ch150 & 6528 & 48726 & 646.415 & 10 & 67.440 & 67.322 & 992.652 \\
 & d198 & 15780 & 158386 & 903.714 & 18 & 160.992 & 232.834 & 2529.323 \\
 & eil101 & 629 & 2813 & 347.218 & 16 & 78.112 & 139.750 & 1630.725 \\
 & kroA100 & 21282 & 135707 & 537.661 & 26 & 129.116 & 272.577 & 3482.952 \\
 & kroA150 & 26524 & 218343 & 723.190 & 6 & 41.580 & 80.523 & 1077.787 \\
 & kroB100 & 22141 & 118419 & 434.840 & 35 & 174.020 & 412.730 & 3539.784 \\
 & kroB150 & 26130 & 202969 & 676.766 & 24 & 166.296 & 200.905 & 3093.957 \\
 & kroC100 & 20749 & 138488 & 567.444 & 16 & 79.440 & 174.127 & 2367.685 \\
 & kroD100 & 21294 & 129635 & 508.787 & 17 & 84.303 & 154.815 & 2220.916 \\
 & kroE100 & 22068 & 125387 & 468.185 & 35 & 174.020 & 413.782 & 3576.009 \\
 & lin105 & 14379 & 102517 & 612.963 & 9 & 46.071 & 97.868 & 1092.099 \\
 & pr107 & 44303 & 303855 & 585.856 & 22 & 117.128 & 285.457 & 3494.023 \\
 & pr124 & 59030 & 526278 & 791.543 & 19 & 114.076 & 213.121 & 3426.667 \\
 & pr136 & 96772 & 713073 & 636.859 & 14 & 90.776 & 163.846 & 1628.570 \\
 & pr144 & 58537 & 118820 & 102.983 & 23 & 156.492 & 350.591 & 3294.159 \\
 & pr152 & 73682 & 849178 & 1052.490 & 23 & 163.864 & 319.237 & 3442.378 \\
 & rat195 & 2323 & 19502 & 739.518 & 21 & 177.534 & 294.969 & 3588.559 \\
 & rd100 & 7910 & 44201 & 458.799 & 19 & 92.036 & 222.474 & 3590.330 \\
 & u159 & 42080 & 347796 & 726.511 & 21 & 155.484 & 294.184 & 3547.076 \\
\midrule
\multirow{15}{*}{0.2-0.5} & a280 & 2579 & 30832 & 1095.502 & 19 & 221.996 & 172.781 & 2004.626 \\
 & d493 & 35002 & 417749 & 1093.500 & 17 & 352.784 & 161.899 & 1966.156 \\
 & fl417 & 11861 & 445256 & 3653.950 & 10 & 173.470 & 84.903 & 1325.056 \\
 & gil262 & 2378 & 24519 & 931.077 & 8 & 89.032 & 79.375 & 1085.446 \\
 & kroA200 & 29368 & 305520 & 940.316 & 16 & 142.192 & 168.055 & 2340.587 \\
 & kroB200 & 29437 & 289238 & 882.566 & 26 & 231.062 & 297.838 & 3578.618 \\
 & lin318 & 41345 & 544122 & 1194.635 & 8 & 108.504 & 89.533 & 837.863 \\
 & pcb442 & 50778 & 723164 & 1324.168 & 8 & 147.848 & 56.720 & 606.033 \\
 & pr226 & 80369 & 1442740 & 1695.145 & 21 & 212.100 & 198.219 & 1902.708 \\
 & pr264 & 49135 & 999930 & 1935.067 & 14 & 162.673 & 109.308 & 1553.069 \\
 & pr299 & 48191 & 677768 & 1306.420 & 23 & 299.092 & 272.404 & 3404.811 \\
 & pr439 & 107217 & 1781733 & 1561.801 & 8 & 148.960 & 96.041 & 1393.436 \\
 & rd400 & 15281 & 195457 & 1179.085 & 21 & 341.124 & 254.880 & 3520.224 \\
 & ts225 & 126643 & 1339414 & 957.630 & 33 & 331.980 & 428.714 & 3322.474 \\
 & tsp225 & 3919 & 30723 & 683.950 & 24 & 230.256 & 281.536 & 3326.143 \\
\midrule
\multirow{6}{*}{0.5-1} & d657 & 48912 & 816654 & 1569.639 & 8 & 218.376 & 78.842 & 810.460 \\
 & p654 & 34643 & 1892175 & 5361.926 & 8 & 217.760 & 82.629 & 1254.455 \\
 & rat575 & 6773 & 106143 & 1467.149 & 9 & 206.046 & 64.595 & 544.923 \\
 & rat783 & 8806 & 167368 & 1800.613 & 11 & 338.778 & 101.534 & 1243.282 \\
 & u574 & 36905 & 639997 & 1634.174 & 13 & 309.712 & 116.383 & 1127.479 \\
 & u724 & 41910 & 814147 & 1842.608 & 13 & 387.374 & 125.961 & 1666.035 \\
\midrule
\multirow{15}{*}{1-2} & d1291 & 50801 & 1659285 & 3166.245 & 9 & 515.853 & 129.673 & 1757.716 \\
 & d1655 & 62128 & 2123203 & 3317.466 & 10 & 782.590 & 106.926 & 976.751 \\
 & fl1400 & 20127 & 1606464 & 7881.637 & 10 & 625.770 & 106.716 & 762.021 \\
 & fl1577 & 22249 & 1310261 & 5789.078 & 8 & 584.680 & 125.307 & 1586.939 \\
 & nrw1379 & 56638 & 1370252 & 2319.316 & 10 & 626.630 & 92.113 & 1529.937 \\
 & pcb1173 & 56892 & 1362385 & 2294.686 & 8 & 402.384 & 114.889 & 1374.360 \\
 & pr1002 & 259045 & 6148455 & 2273.508 & 12 & 494.088 & 140.352 & 1893.208 \\
 & rl1304 & 252948 & 8924001 & 3427.998 & 12 & 700.392 & 147.164 & 1848.956 \\
 & rl1323 & 270199 & 7697390 & 2748.786 & 8 & 474.632 & 96.158 & 864.331 \\
 & rl1889 & 316536 & 14023995 & 4330.458 & 16 & 1466.704 & 214.204 & 2081.738 \\
 & u1060 & 224094 & 6529359 & 2813.670 & 8 & 355.840 & 76.381 & 728.351 \\
 & u1432 & 152970 & 3772731 & 2366.321 & 15 & 985.260 & 184.622 & 2175.513 \\
 & u1817 & 57201 & 2040136 & 3466.609 & 10 & 870.070 & 133.765 & 1910.557 \\
 & vm1084 & 239297 & 8130080 & 3297.485 & 13 & 594.997 & 168.779 & 1424.989 \\
 & vm1748 & 336556 & 14404027 & 4179.831 & 17 & 1421.727 & 220.912 & 2143.461 \\
\midrule
\multirow{7}{*}{2-5} & d2103 & 80450 & 3136167 & 3798.281 & 8 & 828.416 & 95.846 & 974.011 \\
 & fl3795 & 28772 & inf & inf & - & - & - & - \\
 & fnl4461 & 152566 & inf & inf & - & - & - & - \\
 & pcb3038 & 137694 & inf & inf & - & - & - & - \\
 & pr2392 & 378032 & 14868838 & 3833.222 & 9 & 1083.636 & 62.958 & 893.720 \\
 & u2152 & 64253 & 2431659 & 3684.507 & 9 & 953.784 & 129.823 & 1924.544 \\
 & u2319 & 234256 & 5867710 & 2404.828 & 9 & 1046.187 & 79.161 & 1211.242 \\
\midrule
\multirow{2}{*}{5-10}  & rl5915 & 655530 & inf & inf & - & - & - & - \\
 & rl5934 & 556045 & inf & inf & - & - & - & - \\
\midrule
\multirow{5}{*}{$\ge$10} & brd14051 & 469445 & inf & inf & - & - & - & - \\
 & d15112 & 1573152 & inf & inf & - & - & - & - \\
 & d18512 & 645488 & inf & inf & - & - & - & - \\
 & rl11849 & 923368 & inf & inf & - & - & - & - \\
 & usa13509 & 19982889 & inf & inf & - & - & - & - \\
\bottomrule
\end{tabular}
\end{table*}

\begin{table*}[h]
\centering
\renewcommand{\arraystretch}{0.55}
\small
\caption{Result of OPRO on the TSPLIB \texttt{EUC\_2D} subset.}
\label{tab:tsp_opro}
\begin{tabular}{crrrrrrrr}
\toprule
Size (k) & Name & Lower bound & Objective value & Gap (\%) & \#API calls & Input tokens (k) & Output tokens (k) & Running time (sec.) \\
\midrule
\multirow{6}{*}{$<$0.1} & berlin52 & 7542 & 14774 & 95.899 & 9 & 12.420 & 2.428 & 61.642 \\
 & eil51 & 426 & 730 & 71.369 & 12 & 17.760 & 3.098 & 111.630 \\
 & eil76 & 538 & 1648 & 206.376 & 9 & 15.297 & 2.985 & 68.994 \\
 & pr76 & 108159 & 277854 & 156.895 & 15 & 32.916 & 4.802 & 157.505 \\
 & rat99 & 1211 & 3207 & 164.895 & 7 & 15.241 & 2.578 & 65.869 \\
 & st70 & 675 & inf & inf & - & - & - & - \\
\midrule
\multirow{21}{*}{0.1-0.2} & bier127 & 118282 &inf & inf  & - & - & - & - \\
 & ch130 & 6110 &inf & inf  & - & - & - & - \\
 & ch150 & 6528 &inf & inf  & - & - & - & - \\
 & d198 & 15780 & 22512 & 42.664 & 9 & 41.007 & 4.905 & 65.386 \\
 & eil101 & 629 &inf & inf  & - & - & - & - \\
 & kroA100 & 21282 &inf & inf  & - & - & - & - \\
 & kroA150 & 26524 &inf & inf  & - & - & - & - \\
 & kroB100 & 22141 & 157184 & 609.926 & 12 & 27.366 & 4.655 & 75.213 \\
 & kroB150 & 26130 & 273236 & 945.681 & 9 & 30.504 & 4.582 & 128.892 \\
 & kroC100 & 20749 &inf & inf  & - & - & - & - \\
 & kroD100 & 21294 & 127458 & 498.564 & 27 & 72.300 & 9.721 & 267.155 \\
 & kroE100 & 22068 & 162325 & 635.569 & 9 & 21.474 & 3.301 & 96.389 \\
 & lin105 & 14379 &inf & inf  & - & -& - &- \\
 & pr107 & 44303 & 74143 & 67.354 & 9 & 22.896 & 3.671 & 73.170 \\
 & pr124 & 59030 &inf & inf  & - & -& - & -\\
 & pr136 & 96772 &inf & inf  & - & - & - & - \\
 & pr144 & 58537 &inf & inf  & -& - &- & - \\
 & pr152 & 73682 & 160979 & 118.479 & 9 & 31.986 & 4.099 & 119.534 \\
 & rat195 & 2323 &inf & inf  & - & - & - & - \\
 & rd100 & 7910 &inf & inf  & - & - &- & - \\
 & u159 & 42080 & 43375 & 3.079 & 9 & 33.348 & 4.138 & 77.461 \\
\midrule
\multirow{15}{*}{0.2-0.5} & a280 & 2579 & 2818 & 9.291 & 9 & 52.629 & 6.444 & 126.892 \\
 & d493 & 35002 &inf & inf  & - & - & - & - \\
 & fl417 & 11861 &inf & inf  & -& -& - & - \\
 & gil262 & 2378 &inf & inf  & - & - & -& - \\
 & kroA200 & 29368 & 328171 & 1017.444 & 7 & 29.741 & 4.716 & 80.113 \\
 & kroB200 & 29437 & 327452 & 1012.384 & 9 & 40.176 & 5.487 & 164.678 \\
 & lin318 & 41345 & 119866 & 185.200 & 7 & 48.956 & 4.542 & 83.120 \\
 & pcb442 & 50778 & 542583 & 968.541 & 9 & 88.044 & 7.438 & 129.833 \\
 & pr226 & 80369 & 110416 & 37.386 & 9 & 46.860 & 6.380 & 103.985 \\
 & pr264 & 49135 & 77223 & 57.166 & 12 & 76.437 & 8.611 & 139.311 \\
 & pr299 & 48191 & 83507 & 73.285 & 18 & 121.149 & 13.592 & 289.296 \\
 & pr439 & 107217 &inf & inf  & - & - & - & - \\
 & rd400 & 15281 & 215556 & 1310.617 & 9 & 74.589 & 6.980 & 173.999 \\
 & ts225 & 126643 & 1484736 & 1072.380 & 15 & 81.471 & 9.050 & 145.865 \\
 & tsp225 & 3919 & inf & inf  & - & - & - & - \\
\midrule
\multirow{6}{*}{0.5-1} & d657 & 48912 & inf & inf  & - & - & - & - \\
 & p654 & 34643 & inf & inf  & - & - & - & -  \\
 & rat575 & 6773 & inf & inf  & - & - & - & - \\
 & rat783 & 8806& inf & inf  & - & - & - & -  \\
 & u574 & 36905 & inf & inf  & - & - & - & - \\
 & u724 & 41910 & inf & inf  & - & - & - & -  \\
\midrule
\multirow{15}{*}{1-2} & d1291 & 50801 & inf & inf  & - & - & - & - \\
 & d1655 & 62128 & inf & inf  & - & - & - & -  \\
 & fl1400 & 20127 & inf & inf  & - & - & - & -  \\
 & fl1577 & 22249 & inf & inf  & - & - & - & -  \\
 & nrw1379 & 56638 & inf & inf  & - & - & - & -  \\
 & pcb1173 & 56892 & inf & inf  & - & - & - & -  \\
 & pr1002 & 259045 & inf & inf  & - & - & - & - \\
 & rl1304 & 252948 & inf & inf  & - & - & - & -  \\
 & rl1323 & 270199 & inf & inf  & - & - & - & -  \\
 & rl1889 & 316536 & inf & inf  & - & - & - & -  \\
 & u1060 & 224094 & inf & inf  & - & - & - & -  \\
 & u1432 & 152970 & inf & inf  & - & - & - & -  \\
 & u1817 & 57201 & inf & inf  & - & - & - & -  \\
 & vm1084 & 239297 & inf & inf  & - & - & - & -  \\
 & vm1748 & 336556 & inf & inf  & - & - & - & - \\
\midrule
\multirow{7}{*}{2-5} & d2103 & 80450 & inf & inf  & - & - & - & - \\
 & fl3795 & 28772 & inf & inf & - & - & - & - \\
 & fnl4461 & 152566 & inf & inf & - & - & - & - \\
 & pcb3038 & 137694 & inf & inf & - & - & - & - \\
 & pr2392 & 378032 & inf & inf  & - & - & - & -  \\
 & u2152 & 64253 & inf & inf  & - & - & - & -  \\
 & u2319 & 234256 & inf & inf  & - & - & - & -  \\
\midrule
\multirow{2}{*}{5-10}  & rl5915 & 655530 & inf & inf & - & - & - & - \\
 & rl5934 & 556045 & inf & inf & - & - & - & - \\
\midrule
\multirow{5}{*}{$\ge$10} & brd14051 & 469445 & inf & inf & - & - & - & - \\
 & d15112 & 1573152 & inf & inf & - & - & - & - \\
 & d18512 & 645488 & inf & inf & - & - & - & - \\
 & rl11849 & 923368 & inf & inf & - & - & - & - \\
 & usa13509 & 19982889 & inf & inf & - & - & - & - \\
\bottomrule
\end{tabular}
\end{table*}

\begin{table*}[h]
\renewcommand{\arraystretch}{0.55}
\centering
\small
\caption{Result of ReEvo(c) on the TSPLIB \texttt{EUC\_2D} subset.}
\label{tab:tsp_reevoc}
\begin{tabular}{crrrrr}
\toprule
Size (k) & Name & Lower bound & Objective value & Gap (\%) & Inference time (sec.) \\
\midrule
\multirow{6}{*}{$<$0.1} & berlin52 & 7542 & 8608 & 14.140 & 0.034 \\
 & eil51 & 426 & 453 & 6.470 & 0.133 \\
 & eil76 & 538 & 579 & 7.620 & 0.190 \\
 & pr76 & 108159 & 120153 & 11.090 & 0.214 \\
 & rat99 & 1211 & 1361 & 12.410 & 0.199 \\
 & st70 & 675 & 773 & 14.540 & 0.183 \\
\midrule
\multirow{21}{*}{0.1-0.2} & bier127 & 118282 & 131049 & 10.790 & 0.521 \\
 & ch130 & 6110 & 6684 & 9.400 & 0.573 \\
 & ch150 & 6528 & 7234 & 10.820 & 0.705 \\
 & d198 & 15780 & 19247 & 21.980 & 1.614 \\
 & eil101 & 629 & 690 & 9.840 & 0.305 \\
 & kroA100 & 21282 & 22952 & 7.850 & 0.310 \\
 & kroA150 & 26524 & 29605 & 11.620 & 0.812 \\
 & kroB100 & 22141 & 24842 & 12.200 & 0.346 \\
 & kroB150 & 26130 & 29532 & 13.020 & 0.812 \\
 & kroC100 & 20749 & 24043 & 15.880 & 0.311 \\
 & kroD100 & 21294 & 23910 & 12.290 & 0.341 \\
 & kroE100 & 22068 & 23967 & 8.610 & 0.330 \\
 & lin105 & 14379 & 15215 & 5.820 & 0.337 \\
 & pr107 & 44303 & 47040 & 6.180 & 0.350 \\
 & pr124 & 59030 & 68244 & 15.610 & 0.568 \\
 & pr136 & 96772 & 108780 & 12.410 & 0.612 \\
 & pr144 & 58537 & 65538 & 11.960 & 0.704 \\
 & pr152 & 73682 & 83779 & 13.700 & 0.822 \\
 & rat195 & 2323 & 2481 & 6.820 & 1.554 \\
 & rd100 & 7910 & 8665 & 9.550 & 0.208 \\
 & u159 & 42080 & 46487 & 10.470 & 0.946 \\
\midrule
\multirow{15}{*}{0.2-0.5} & a280 & 2579 & 3067 & 18.930 & 4.387 \\
 & d493 & 35002 & 39701 & 13.430 & 22.934 \\
 & fl417 & 11861 & 14132 & 19.150 & 14.108 \\
 & gil262 & 2378 & 2685 & 12.940 & 3.699 \\
 & kroA200 & 29368 & 32620 & 11.080 & 1.635 \\
 & kroB200 & 29437 & 35024 & 18.980 & 1.687 \\
 & lin318 & 41345 & 49017 & 16.630 & 6.110 \\
 & pcb442 & 50778 & 57607 & 13.450 & 16.612 \\
 & pr226 & 80369 & 94848 & 18.020 & 2.334 \\
 & pr264 & 49135 & 57378 & 16.780 & 3.528 \\
 & pr299 & 48191 & 58131 & 20.630 & 5.305 \\
 & pr439 & 107217 & 127860 & 19.250 & 16.301 \\
 & rd400 & 15281 & 17255 & 12.920 & 12.757 \\
 & ts225 & 126643 & 134946 & 6.560 & 2.236 \\
 & tsp225 & 3919 & 4350 & 11.020 & 2.319 \\
\midrule
\multirow{6}{*}{0.5-1} & d657 & 48912 & 56758 & 16.040 & 54.088 \\
 & p654 & 34643 & 41121 & 18.700 & 53.232 \\
 & rat575 & 6773 & 7638 & 12.780 & 36.041 \\
 & rat783 & 8806 & 10018 & 13.770 & 89.825 \\
 & u574 & 36905 & 44434 & 20.400 & 35.946 \\
 & u724 & 41910 & 48979 & 16.870 & 72.058 \\
\midrule
\multirow{15}{*}{1-2} & d1291 & 50801 & 58731 & 15.610 & 402.530 \\
 & d1655 & 62128 & 73008 & 17.510 & 850.869 \\
 & fl1400 & 20127 & 24096 & 19.720 & 494.310 \\
 & fl1577 & 22249 & 24944 & 12.110 & 734.272 \\
 & nrw1379 & 56638 & 64780 & 14.380 & 497.545 \\
 & pcb1173 & 56892 & 66840 & 17.490 & 308.277 \\
 & pr1002 & 259045 & 310008 & 19.670 & 189.077 \\
 & rl1304 & 252948 & 303725 & 20.070 & 421.773 \\
 & rl1323 & 270199 & 317559 & 17.530 & 439.428 \\
 & rl1889 & 316536 & 371919 & 17.500 & 1279.345 \\
 & u1060 & 224094 & 266469 & 18.910 & 223.714 \\
 & u1432 & 152970 & 172960 & 13.070 & 547.619 \\
 & u1817 & 57201 & 66718 & 16.640 & 1120.653 \\
 & vm1084 & 239297 & 280617 & 17.270 & 245.172 \\
 & vm1748 & 336556 & 396194 & 17.720 & 1024.559 \\
\midrule
\multirow{7}{*}{2-5} & d2103 & 80450 & 86631 & 7.680 & 1763.146 \\
 & fl3795 & 28772 &inf & inf  & 3600.101 \\
 & fnl4461 & 152566 &inf & inf  & 3600.103 \\
 & pcb3038 & 137694 &inf & inf  & 3600.029 \\
 & pr2392 & 378032 & 459589 & 21.570 & 2617.235 \\
 & u2152 & 64253 & 76497 & 19.060 & 1874.302 \\
 & u2319 & 234256 & 248857 & 6.230 & 2318.946 \\
\midrule
\multirow{2}{*}{5-10}& rl5915 & 655530 &inf & inf  & 39401.084 \\
 & rl5934 & 556045 &inf & inf  & 3600.103 \\
\midrule
\multirow{5}{*}{$\ge$10} & brd14051 & 469445 &inf & inf  & 3600.107 \\
 & d15112 & 1573152 &inf & inf  & 3600.107 \\
 & d18512 & 645488 &inf & inf  & 3600.109 \\
 & rl11849 & 923368 &inf & inf  & 3600.096 \\
 & usa13509 & 19982889 &inf & inf  & 3600.106 \\
\bottomrule
\end{tabular}
\end{table*}

\begin{table*}[h]
\renewcommand{\arraystretch}{0.55}
\centering
\small
\caption{Result of ReEvo(a) on the TSPLIB \texttt{EUC\_2D} subset.}
\label{tab:tsp_reevoa}
\begin{tabular}{crrrrr}
\toprule
Size (k) & Name & Lower bound & Objective value & Gap (\%) & Inference time (sec.) \\
\midrule
\multirow{6}{*}{$<$0.1} & berlin52 & 7542 & 7549 & 0.100 & 1.873 \\
 & eil51 & 426 & 444 & 4.410 & 1.199 \\
 & eil76 & 538 & 561 & 4.310 & 1.276 \\
 & pr76 & 108159 & 116960 & 8.140 & 3.068 \\
 & rat99 & 1211 & 1287 & 6.350 & 3.670 \\
 & st70 & 675 & 718 & 6.370 & 1.174 \\
\midrule
\multirow{21}{*}{0.1-0.2} & bier127 & 118282 & 125723 & 6.290 & 2.270 \\
 & ch130 & 6110 & 6483 & 6.110 & 2.302 \\
 & ch150 & 6528 & 6767 & 3.670 & 6.842 \\
 & d198 & 15780 & 17404 & 10.290 & 3.839 \\
 & eil101 & 629 & 669 & 6.380 & 1.743 \\
 & kroA100 & 21282 & 22599 & 6.190 & 1.715 \\
 & kroA150 & 26524 & 29047 & 9.510 & 2.709 \\
 & kroB100 & 22141 & 23370 & 5.550 & 1.712 \\
 & kroB150 & 26130 & 29227 & 11.860 & 2.752 \\
 & kroC100 & 20749 & 21599 & 4.100 & 1.706 \\
 & kroD100 & 21294 & 22831 & 7.220 & 1.703 \\
 & kroE100 & 22068 & 23727 & 7.520 & 1.704 \\
 & lin105 & 14379 & 14954 & 4.000 & 1.818 \\
 & pr107 & 44303 & 47092 & 6.300 & 1.874 \\
 & pr124 & 59030 & 61293 & 3.830 & 2.217 \\
 & pr136 & 96772 & 108748 & 12.380 & 2.419 \\
 & pr144 & 58537 & 60957 & 4.140 & 6.187 \\
 & pr152 & 73682 & 76744 & 4.160 & 5.508 \\
 & rat195 & 2323 & 2481 & 6.810 & 3.734 \\
 & rd100 & 7910 & 8485 & 7.280 & 3.537 \\
 & u159 & 42080 & 44710 & 6.250 & 5.823 \\
\midrule
\multirow{15}{*}{0.2-0.5} & a280 & 2579 & 2946 & 14.260 & 7.071 \\
 & d493 & 35002 & 38921 & 11.200 & 12.795 \\
 & fl417 & 11861 & 13568 & 14.400 & 9.877 \\
 & gil262 & 2378 & 2618 & 10.100 & 10.935 \\
 & kroA200 & 29368 & 32150 & 9.470 & 8.430 \\
 & kroB200 & 29437 & 32924 & 11.850 & 8.610 \\
 & lin318 & 41345 & 47067 & 11.990 & 17.050 \\
 & lin318 & 42029 & 47067 & 11.990 & 17.050 \\
 & pcb442 & 50778 & 59216 & 16.620 & 25.854 \\
 & pr226 & 80369 & 89105 & 10.870 & 4.472 \\
 & pr264 & 49135 & 54196 & 10.300 & 5.357 \\
 & pr299 & 48191 & 53895 & 11.840 & 6.433 \\
 & pr439 & 107217 & 119854 & 11.790 & 22.855 \\
 & rd400 & 15281 & 17483 & 14.410 & 23.821 \\
 & ts225 & 126643 & 132257 & 4.430 & 4.455 \\
 & tsp225 & 3919 & 4306 & 9.880 & 9.928 \\
\midrule
\multirow{6}{*}{0.5-1} & d657 & 48912 & 56706 & 15.940 & 19.410 \\
 & p654 & 34643 & 41239 & 19.040 & 40.870 \\
 & rat575 & 6773 & 7789 & 15.000 & 15.848 \\
 & rat783 & 8806 & 10293 & 16.890 & 25.583 \\
 & u574 & 36905 & 43382 & 17.550 & 15.757 \\
 & u724 & 41910 & 48290 & 15.220 & 22.880 \\
\midrule
\multirow{15}{*}{1-2} & d1291 & 50801 & 58520 & 15.200 & 70.189 \\
 & d1655 & 62128 & 73286 & 17.960 & 161.547 \\
 & fl1400 & 20127 & 24131 & 19.900 & 50.081 \\
 & fl1577 & 22249 & 25637 & 15.230 & 134.016 \\
 & nrw1379 & 56638 & 67191 & 18.630 & 52.296 \\
 & pcb1173 & 56892 & 68553 & 20.500 & 91.328 \\
 & pr1002 & 259045 & 303407 & 17.130 & 80.787 \\
 & rl1304 & 252948 & 302084 & 19.430 & 94.308 \\
 & rl1323 & 270199 & 316894 & 17.280 & 48.574 \\
 & rl1889 & 316536 & 375685 & 18.690 & 86.597 \\
 & u1060 & 224094 & 269036 & 20.060 & 41.572 \\
 & u1432 & 152970 & 182777 & 19.490 & 107.057 \\
 & u1817 & 57201 & 65498 & 14.510 & 199.884 \\
 & vm1084 & 239297 & 283420 & 18.440 & 111.560 \\
 & vm1748 & 336556 & 406986 & 20.930 & 171.830 \\
\midrule
\multirow{7}{*}{2-5} & d2103 & 80450 & 85620 & 6.430 & 103.428 \\
 & fl3795 & 28772 & 34390 & 19.530 & 212.787 \\
 & fnl4461 & 152566 & 221155 & 44.960 & 259.595 \\
 & pcb3038 & 137694 & 166550 & 20.960 & 159.196 \\
 & pr2392 & 378032 & 458366 & 21.250 & 247.613 \\
 & u2152 & 64253 & 76332 & 18.800 & 100.445 \\
 & u2319 & 234256 & 272864 & 16.480 & 103.202 \\
\midrule
\multirow{2}{*}{5-10} & rl5915 & 655530 & 686839 & 4.780 & 1019.255 \\
 & rl5934 & 556045 & 674165 & 21.240 & 473.464 \\
\midrule
\multirow{5}{*}{$\ge$10} & brd14051 & 469445 & 577012 & 22.910 & 6294.430 \\
 & d15112 & 1573152 & 1943279 & 23.530 & 7639.086 \\
 & d18512 & 645488 & 794641 & 23.110 & 6152.544 \\
 & rl11849 & 923368 & 1130865 & 22.470 & 2086.743 \\
 & usa13509 & 19982889 & 25269411 & 26.460 & 3008.846 \\
\bottomrule
\end{tabular}
\end{table*}



\begin{table*}[h]
\centering
\small
\caption{Result of \textbf{DRAGON} on the CVRPLIB (X/XML) subset.}
\label{tab:cvrp_llmda}
\begin{tabular}{crrrrrrrr}
\toprule
Size (k) & Name & Lower bound & Objective value & Gap (\%) & \#API calls & Input tokens (k) & Output tokens (k) & Running time (sec.) \\
\midrule
\multirow{2}{*}{0.1-0.2} & X-n110-k13 & 14971 & 19298 & 28.903 & 16 & 235.218 & 27.996 & 974.263 \\
 & X-n162-k11 & 14138 & 17280 & 22.224 & 15 & 330.999 & 31.953 & 792.177 \\
\midrule
\multirow{4}{*}{0.2-0.5} & X-n223-k34 & 40437 & 52877 & 30.764 & 13 & 249.044 & 23.135 & 710.500 \\
 & X-n284-k15 & 20226 & 26916& 33.076 & 31 & 754.018 & 64.900 & 2080.547 \\
 & X-n420-k130 & 107798 & 133902 & 24.216 & 52 & 1238.116 & 110.004 & 3557.632 \\
 & X-n459-k26 & 24139 & 31597 & 30.896 & 32 & 1088.167 & 80.217 & 2263.910 \\
\midrule
\multirow{5}{*}{0.5-1} 
 & X-n513-k21 & 24201 & 31918 & 31.887 & 39 & 1445.657 & 84.518 & 2477.514 \\
 & X-n670-k130 & 146332 & 211859 & 44.780 & 24 & 1181.644 & 75.815 & 2227.067 \\
 & X-n701-k44 & 81923 & 94156 & 14.932 & 6 & 251.993 & 15.387 & 454.379 \\
 & X-n783-k48 & 72386 & 94986 & 31.222 & 30 & 1117.956 & 88.571 & 2530.259 \\
 & X-n801-k40 & 73311 & 81258 & 10.840 & 26 & 971.698 & 53.939 & 1820.531 \\
\midrule
\multirow{3}{*}{1-2} 
& X-n1001-k43 & 72355 & 85873 & 18.683 & 48 & 1985.763 & 125.876 & 3510.914 \\
& XML1000\_1335\_01 & 63968 & 73845 &  15.441 & 16 & 489.379 & 27.625 & 825.463 \\
& XML1500\_2114\_01 & 99781 & 112163 & 12.409 & 24 & 1855.369 & 101.138 & 2936.169 \\
\midrule
\multirow{3}{*}{2-5} 
& XML2500\_2145\_01 & 112858 & 154727 & 37.099 & 23 & 1557.874 & 76.582 & 1948.524 \\
& XML3500\_3335\_01 & 284329 & 300053 & 5.530  & 49 & 3397.328 & 161.908 & 3482.792 \\
& XML4500\_3233\_01 & 602565 & 625500 & 3.806 & 32 & 3373.244 & 150.978 & 3480.450 \\
\midrule
\multirow{3}{*}{$\ge$5}
& XML5000\_2224\_01 & 315739 & 337855 & 7.005 & 28 & 5393.904 & 204.343 & 3549.789 \\
& XML5000\_1321\_01 & 1466910 & 1566502 & 6.789 & 21 & 3137.194 & 91.763 & 2735.975 \\
& XML5000\_3135\_01 & 396487 & 418520 & 5.557 & 26 & 2331.798 & 111.974 & 3028.920 \\
\bottomrule
\end{tabular}
\end{table*}

\begin{table*}[h]
\centering
\caption{Result of ReEvo(a) on the  CVRPLIB (X/XML) subset.}
\label{tab:cvrp_reevo}
\small
\begin{tabular}{crrrrr}
\toprule
Size (k) & Name & Lower bound & Objective value & Gap (\%) & Inference time (sec.) \\
\midrule
\multirow{2}{*}{0.1-0.2} & X-n110-k13 & 14971 & 17758.55 & 18.620 & 3.293  \\
 & X-n162-k11 & 14138 & 17581.21 & 24.350 & 5.086 \\
\midrule
\multirow{4}{*}{0.2-0.5} & X-n223-k34 & 40437 & 47539.29 & 17.560 & 8.326 \\
 & X-n284-k15 & 20226 & 25702.50 & 27.080 & 11.032 \\
 & X-n420-k130 & 107798 & 122008.00 & 13.180 & 22.199 \\
 & X-n459-k26 & 24139 & 32910.35 & 36.340 & 21.046 \\
\midrule
\multirow{5}{*}{0.5-1} 
 & X-n513-k21 & 24201 & 34441.10 & 42.310 & 24.540 \\
 & X-n670-k130 & 146332 & 179082.02 & 22.380 & 43.652 \\
 & X-n701-k44 & 81923 & 99286.02 & 21.190 & 40.841 \\
 & X-n783-k48 & 72386 & 101879.89 & 40.750 & 49.255 \\
 & X-n801-k40 & 73311 & 87669.11 & 19.590 & 49.812 \\
\midrule
\multirow{3}{*}{1-2} 
& X-n1001-k43 & 72335 & 94073.34 & 30.020 & 72.013 \\
& XML1000\_1335\_01 & 63968 & 80300.27 & 25.530 & 69.735 \\
& XML1500\_2114\_01 & 99781 & 120350.96  & 20.620  & 101.063 \\
\midrule
\multirow{3}{*}{2-5}
& XML2500\_2145\_01 & 112858 & 164532.37 & 45.790 & 189.593 \\
& XML3500\_3335\_01 & 284329 & 322280.75 & 13.350 & 289.669 \\
& XML4500\_3233\_01 & 602565 & 643929.79 & 6.860 & 418.924\\
\midrule
\multirow{3}{*}{$\ge$5}
& XML5000\_2224\_01 & 315739 & 359751.70 & 13.940 & 489.946 \\
& XML5000\_1321\_01 & 1466910 & 1570524.76 & 7.063 & 567.689 \\
& XML5000\_3135\_01 & 396487 & 446867.32 & 12.710 & 470.686 \\
\bottomrule
\end{tabular}
\end{table*}

\begin{table*}[h]
    \centering
    \caption{Ablation for Decomposition and Reconstruction strategies}
    \label{tab:ablation_strategy}
    \small
    \begin{tabular}{cccrrrr}
    \toprule
    Decomposition & Reconstruction & Name & Gap (\%) & Input tokens (k) & Output tokens (k) & Time (sec)\\
    \midrule
    \multirow{15}{*}{Random}   & \multirow{5}{*}{Heuristic} 
                            & eil51 & 5.634 & - & - & 0.508  \\
                &           & pr299 & 12.83 & - & - & 1.043 \\
                &           & u1060 & 12.166 & - & - & 3.013 \\
                &           & brd14051 & 13.621 & - & - & 124.368 \\\cmidrule(lr){3-7}
                &           & Average & 10.938 & - & - & 32.233 \\\cmidrule(lr){2-7}
                & \multirow{5}{*}{Solver} & eil51 & 5.634 & - & - & 1.118 \\
                &           & pr299 & 8.688	& - & - & 601.031 \\
                &           & u1060 & 12.374 & - & - & 602.940 \\
                &           & brd14051 & 13.386 & - & - &724.978 \\\cmidrule(lr){3-7}
                &           & Average & 10.021 & - & - & 482.517 \\\cmidrule(lr){2-7}
                & \multirow{5}{*}{LLM} & eil51 & 7.981 & 66.143 & 10.258 & 101.244 \\
                &           & pr299 & 13.851 & 93.981 & 14.448 & 821.860 \\
                &           & u1060 & 11.173 & 89.304 & 11.820 & 656.243 \\
                &           & brd14051 & 13.196 & 276.588 & 25.373 & 1520.241 \\\cmidrule(lr){3-7}
                &           & Average &  11.550 & 131.504 & 15.475 & 774.870 \\
    \midrule
    \multirow{15}{*}{Heuristic}   & \multirow{5}{*}{Heuristic} & eil51 & 4.695 & - & - & 0.504 \\
                &           & pr299  & 15.688 & - & - & 1.056 \\
                &           & u1060 & 14.009 & - & - & 3.195 \\
                &           & brd14051 & 13.13 & - & - & 157.806 \\\cmidrule(lr){3-7}
                &           & Average & 11.881 & - & - & 40.640 \\\cmidrule(lr){2-7}
                & \multirow{5}{*}{Solver} & eil51 & 5.399 & - & - & 600.516 \\
                &           & pr299 & 10.207 & - & - & 701.056 \\
                &           & u1060  & 11.194 & - & - & 603.169 \\
                &           & brd14051 & 12.942 & - & - & 753.731 \\\cmidrule(lr){3-7}
                &           & Average & 9.936 & - & - & 664.618 \\\cmidrule(lr){2-7}
                & \multirow{5}{*}{LLM} & eil51 & 5.869 & 249.725 & 29.895 & 1653.207 \\
                &           & pr299 & 10.747 & 105.327 & 16.817 & 833.485 \\
                &           & u1060 & 13.170 & 75.322 & 15.279 & 1653.207 \\
                &           & brd14051 & 13.374 & 104.041 & 19.172 & 954.097 \\\cmidrule(lr){3-7}
                &           & Average & 10.790 & 133.604 & 20.291 & 1273.499 \\
    \midrule
    \multirow{15}{*}{LLM}   & \multirow{5}{*}{Heuristic} & eil51 & 5.634 & 196.966 & 9.220 & 407.764 \\
                &           & pr299 & 11.436 & 318.755 & 11.862 & 458.776 \\
                &           & u1060 & 10.961 & 420.500 & 17.615 & 462.020 \\
                &           & brd14051 & 13.094 & 1389.088 & 3.652 & 1081.917 \\\cmidrule(lr){3-7}
                &           & Average & 10.281 & 581.327 & 10.587 & 602.619 \\\cmidrule(lr){2-7}
                & \multirow{5}{*}{Solver} & eil51  & 5.634 & 188.621 & 10.222 & 535.234 \\
                &           & pr299 & 8.246 & 440.071 & 19.319 & 709.121 \\
                &           & u1060 & 10.961 & 352.112 & 28.451 & 642.717 \\
                &           & brd14051 & 13.079 & 10609.444 & 116.235 & 3260.043 \\\cmidrule(lr){3-7}
                &           & Average & \textbf{9.480} & 2897.562 & 43.557 & 1286.779 \\\cmidrule(lr){2-7}
                & \multirow{5}{*}{LLM} & eil51 & 3.286 & 355.136 & 33.297 & 1775.789 \\
                &           & pr299 & 11.612 & 333.709 & 30.133 & 997.101 \\
                &           & u1060 & 10.961 & 349.302 & 33.205 & 986.517 \\
                &           & brd14051 & 13.094 & 2502.501 & 37.871 & 1950.459 \\\cmidrule(lr){3-7}
                &           & Average & \underline{9.738} & 885.162 & 33.627 & 1427.466 \\
    \bottomrule
    \end{tabular}
\end{table*}

\begin{table*}[h]
\centering
\caption{Impact of different LLM models}
\label{tab:ablation_llm}
\small
\begin{tabular}{crrrrrrr}\toprule
LLM model & Name & Gap (\%) & \#API calls & Input tokens (k) & Output tokens (k) & Running time (sec.) \\\midrule
\multirow{6}{*}{\texttt{gpt-4o}} &berlin52 &16.322 &18 & 352.398 &27.373 &1153.372 \\
&kroB200 &12.637 &40 &803.297 &74.733 &3517.805 \\
&d2103 &24.021 &23 &888.182 &74.305 &2711.948 \\
&rl5934 &22.228 &18 &3465.698 &149.070 &2340.457 \\
&rl11849 &20.078 &12 &1261.444 &36.615 &1398.704 \\\cmidrule(lr){2-7}
& Average & 19.057 & 22.20  & 1354.204 & 72.419 & 2224.457 \\
\midrule
\multirow{6}{*}{\texttt{deepseek-reasoner}} &berlin52 &13.710 &4 &2.226 &48.926 &2431.104 \\
&kroB200 &14.190 &4 &5.031 &45.960 &2423.599 \\
&d2103 &23.052 &2 &22.289 &25.028 &2183.935 \\
&rl5934 &21.466 &4 &129.526 &65.429 &3036.950 \\
&rl11849 & inf & \multicolumn{4}{l}{(Exceeds the model's max input limit)} \\\cmidrule(lr){2-7}
& Average & inf & - & - & - & - \\\midrule
\multirow{6}{*}{\texttt{o3}} &berlin52 &3.248 &20 &10.425 &182.739 &3422.398 \\
&kroB200 &10.147 &12 &15.368 &152.977 &3343.023 \\
&d2103 &22.933 &10 &112.905 &115.582 &3220.190 \\
&rl5934 &21.672 &14 &456.244 &105.146 &2220.981 \\
&rl11849 &20.192 &12 &781.328 &107.737 &2150.865 \\\cmidrule(lr){2-7}
& Average &\textbf{15.638} & 13.60 &275.254 &132.836 &2871.491 \\
\midrule
\multirow{6}{*}{\texttt{gpt-4.1}} &berlin52 &6.457 &12 &7.011 &13.566 &289.156 \\
&kroB200 &8.900 &13 &18.041 &17.039 &284.020 \\
&d2103 &23.693 &12 &135.993 &18.674 &421.396 \\
&rl5934 &22.134 &15 &522.483 &13.861 &349.562 \\
&rl11849 &19.552 &12 &781.658 &15.665 &433.936 \\\cmidrule(lr){2-7}
& Average & \underline{16.147} & 12.80 & 293.037 & 15.761 & 355.614 \\
\bottomrule
\end{tabular}
\end{table*}

\begin{table*}
[ht]
\centering
\small
\setlength{\tabcolsep}{0.9mm}
\caption{Result of \textbf{DRAGON} for Bin Packing Problem.}\label{tab:bpp_llmdr}
\begin{tabular}{lrrrrrrr}\toprule
Name &Objective value & L2 lower bound & Running time (sec.) &Gap (\%)  &\# API calls &Input tokens (k) &Output tokens (k) \\\midrule
Weibull 5k/test\_0 &2020 & 2012&512.607 &0.398  &19 &238.147 &28.797 \\
Weibull 5k/test\_1 &1984 & 1983&527.750 &0.050  &15 &195.512 &22.628 \\
Weibull 5k/test\_2 &1990 & 1978&122.899 &0.607  & 4 &40.045 &7.187 \\
Weibull 5k/test\_3 &1992 & 1986 &513.842 &0.302 & 18 &240.698 &32.893 \\
Weibull 5k/test\_4 &1986 & 1980 &762.268 &0.303 & 17 &235.260 &29.430 \\
\bottomrule
\end{tabular}
\end{table*}

\begin{table*}[ht]
\small
\centering
\caption{Result of ReEvo(a) for Bin Packing Problem.}\label{tab:bpp_reevo}
\begin{tabular}{lrrrr}\toprule
Name &Objective value & L2 lower bound &Running time (sec.) &Gap (\%) \\\midrule
Weibull 5k/test\_0 &2079 & 2012 &57.687 &3.330 \\
Weibull 5k/test\_1 &2052 & 1983 &55.274 &3.480 \\
Weibull 5k/test\_2 &2050 & 1990 &55.965 &3.640 \\
Weibull 5k/test\_3 &2053 & 1992 &57.682 &3.374 \\
Weibull 5k/test\_4 &2049 & 1986 & 56.779 &3.485 \\
\bottomrule
\end{tabular}
\end{table*}

\begin{table*}[ht]
\small
\centering
\setlength{\tabcolsep}{0.8mm}
\caption{Result of \textbf{DRAGON} for MKP.}\label{tab:mkp_llmdr}
\begin{tabular}{lrrrrrrrr}\toprule
Name &Objective value &Upper bound &Gap (\%) &Running time (sec.) &\# API calls &Input tokens (k) &Output tokens (k) \\\midrule
mkp\_393\_5\_0 &5427 &5455 & 0.513 &345.003 &13 &20.439 &12.424 \\
mkp\_565\_10\_0 &11415 &11478 &0.549 &275.914 &12 &27.345 &11.482 \\
mkp\_3594\_20\_0 &37690 &37706 &0.042 &159.304 &12 &143.961 &6.545 \\
mkp\_3851\_20\_0 &37018 &37058 &0.108 &223.993 &13 &153.656 &8.525 \\
mkp\_6870\_50\_0 &83826 &83880 &0.064 &213.504 &12 &275.633 &12.221 \\
mkp\_6922\_50\_0 &80305 &80416 &0.138 &434.103 &19 &282.751 &28.851 \\
mkp\_8864\_50\_0 &85152 &85266 &0.134 &320.866 &16 &350.271 &17.373 \\
mkp\_10729\_80\_0 &127696 &127756 &0.047 &244.321 &15 &430.355 &13.235 \\
mkp\_12397\_50\_0 &105562 &105592 &0.028 &123.530 &7 &240.881 &5.239 \\
mkp\_30018\_100\_0 &235436 &235483 &0.020 &167.235 &12 &1152.107 &3.960 \\
\bottomrule
\end{tabular}
\end{table*}

\begin{table*}[ht]
\small
\centering
\caption{Result of OR-Tools (CP-SAT) for MKP.}\label{tab:mkp_solver}
\begin{tabular}{lrrrrr}\toprule
Name &Objective value &Upper bound &Gap (\%) &Running time (sec.) \\\midrule
mkp\_393\_5\_0 &5455 &5455 &0.000 &2.52 \\
mkp\_565\_10\_0 &11476 &11478 &0.017 &3600.69 \\
mkp\_3594\_20\_0 &37706 &37706 &0.000&80.11 \\
mkp\_3851\_20\_0 &37058 &37058 &0.000&144.89 \\
mkp\_6870\_50\_0 &83872 &83880 &0.010 &3619.34 \\
mkp\_6922\_50\_0 &80415 &80416 &0.001 &3607.14 \\
mkp\_8864\_50\_0 &85261 &85266 &0.006 &3608.45 \\
mkp\_10729\_80\_0 &127723 &127756 &0.026 &3614.47 \\
mkp\_12397\_50\_0 &105584 &105592 &0.008 &3610.41 \\
mkp\_30018\_100\_0 &227828 &235483 &3.251 &3611.22 \\
\bottomrule
\end{tabular}
\end{table*}


\end{document}